\pdfoutput=1

\documentclass[11pt]{article}

\usepackage[]{main}

\usepackage{times}
\usepackage{latexsym}

\usepackage[T1]{fontenc}

\usepackage[utf8]{inputenc}

\usepackage{microtype}

\usepackage{booktabs}  
\usepackage{graphicx}  
\usepackage{xcolor}  
\usepackage{multirow}  
\newcommand{\pwturn}{PW-Turn}
\newcommand{\pwdialog}{PW-Dialog}
\newcommand{\smturn}{SM-Turn}
\newcommand{\smdialog}{SM-Dialog}

\title{
Human Evaluation of Conversations is an Open Problem:\\
{\em comparing the sensitivity of various methods for evaluating dialogue agents}}

\author{Eric Michael Smith$^1$ \quad Orion Hsu$^2$ \quad Rebecca Qian$^1$ \\
\textbf{Stephen Roller$^1$ \quad Y-Lan Boureau$^1$ \quad Jason Weston$^1$} \\
\\
\begin{tabular}{ll}
  $^1$Facebook AI Research \quad & $^2$Duke University \\
\end{tabular}
}

\begin{document}
\maketitle
\begin{abstract}
At the heart of improving conversational AI is the open problem of how to evaluate conversations. 
Issues with automatic metrics are
well known \cite{liu2016not}, with human evaluations still considered the gold standard. Unfortunately, how to perform human evaluations is also an open problem:  differing data collection methods have varying levels of human agreement and statistical sensitivity, 
resulting in differing amounts of human annotation hours and labor costs. In this work we compare five different crowdworker-based human evaluation methods and find that different methods are best depending on the types of models compared, with no clear winner across the board.
While this highlights the open problems in the area, 
our analysis leads to advice of when to use which one, and possible future 
directions. 

\end{abstract}

\section{Introduction}

Any comprehensive analysis of the performance of an open-domain conversational model must include human evaluations: automatic metrics  can capture certain aspects of model performance but are no replacement for having human raters judge how adept models are at realistic and interesting conversation \cite{deriu2021survey,liu2016not,dinan2019second}. Unfortunately, human evaluations themselves must be carefully constructed in order to capture all the aspects desired of a good conversationalist. Any evaluation technique must evaluate over many turns of a conversation in order to detect emergent faults such as repetitiveness or contradiction, while techniques that rely solely on a single evaluation at the end of a conversation may fail to take into account changes in model performance over its span. Further, techniques that rate model performance on a Likert scale may suffer from inconsistencies in subjective numerical ratings across evaluations of different models \citep{li2019acute}. 
When comparing various human evaluation methods to assess which works best, we find that each has success and failure cases, leading us to conclude that human evaluation is still an open problem.


In this work, we analyze a representative set of human evaluation techniques.
First, we compare {\em per-turn} evaluations, where ratings are given
after every model response, and {\em per-dialogue} evaluations, where ratings are collected solely at the end of the conversation. Per-turn evaluations have the advantage of being more fine-grained, encouraging annotators to focus on small differences; however, the quality of a conversation is more than the sum of its parts, and global {per-dialogue} evaluations can capture this better. 
Second, we consider {\em pairwise methods}, where two models are compared directly by an annotator, to {\em single-model} methods, where the annotator sees and rates only one model at a time.  Both approaches can be either per-turn or per-dialogue.
For example, in Pairwise Per-Turn evaluation, a crowdworker chats with a dialogue agent, and after each of the worker's messages, they must choose between two possible responses from the agent, one from each of two different models. 
The pairwise approach can spot subtle differences apparent when comparing responses, and it can mitigate problems with distribution shift that occur in absolute scoring. Single-model approaches, however, can work well when direct comparison is not paramount.

We compare all of these different techniques for evaluating dialogue models in three different settings, and we contrast their individual strengths. We find that:
\begin{itemize}
  \item Pairwise per-turn evaluations are adept at measuring changes in model performance throughout a conversation. 
  This technique tends to work well when pairs of models clearly differ in how appropriate their responses are in the context of the previous lines of dialogue, for example, when comparing two models that are trained on different datasets.
  \item Pairwise per-dialogue evaluations tend to perform best when differences between models only emerge after several conversation turns, such  as when these differences are very subtle, or when noticing patterns in responses that emerge globally across the entire conversation, for example the average length of responses.
  \item Single-model evaluations, performed both per conversation turn and at the end of a conversation, tend to not do as well in the two previously described  settings, but do perform well when comparing models that differ only slightly in quality but are otherwise similar, for example two models with different numbers of parameters.
\end{itemize}

These findings, while highlighting the difficulty of human evaluation, also provide guidance on which method might be best to use in these different circumstances, as well as possible future work. In particular, investigating the best way to merge pairwise and single-model, per-turn and per-dialogue benefits into a single method could be a fruitful direction. We also analyze the interpretability of these approaches when collecting human written explanations. 
We will soon release code for these evaluation techniques in the ParlAI framework.\footnote{\url{https://parl.ai/projects/humaneval}}

\section{Existing work}

\paragraph{Open-domain versus specific domain}
Our work concentrates on the open-domain setting.
In specific conversational domains, such as question answering (QA), evaluation can be simpler and is often reduced to measuring overlap or exact match with the correct answer \cite{chen2019evaluating}. However, this no longer as easily suffices for free-form, conversational and long-form QA where answers are more open-ended \cite{fan2019eli5,adolphs2021reason}. Similarly, for certain types of goal-oriented conversations more targeted evaluations can take place, for example evaluation of state tracking \cite{williams2016dialog} and task completion \citep{lemon,henderson2014second,bordes2016learning,asri2017frames,wen2016network}.  Open-domain dialogue potentially covers all these other cases as special cases, 
but also covers conversations that are more free-form or do not have a precise goal. Hence, finding a reliable evaluation technique is more difficult, and there is currently no single standard method that is agreed upon \cite{deriu2021survey,huang2020challenges,roller2020open}. Different techniques that have been proposed will be described in the following paragraphs.

\paragraph{Automatic metrics}

Automatic metrics are the most convenient for fast, efficient and reproducible research with a quick turn-around and development cycle, hence they are frequently used. 
Unfortunately, many of them, such as BLEU, METEOR and ROUGE
have been shown to only “correlate very weakly with human judgement” \cite{liu2016not}. A central problem is that due to the open-ended nature of conversations, there are many possible responses in a given dialogue, and, while having multiple references can help, there is typically only one gold label available \cite{gupta2019investigating}.
Perplexity (computing the predicted probability of the given gold utterances) has been argued to correlate with human judgments \cite{adiwardana2020towards}, however this has also been shown to not always be the case \cite{dinan2019second}, and moreover does not actually evaluate the generations themselves produced by a decoder architecture. Hence, changing the behavior of the generation method can dramatically change human evaluations, while maintaining identical or near-identical perplexity \citep{see2019goodconversation,welleck2019neuraltext,adiwardana2020towards,roller2021recipes}.
An alternative recent trend is to employ trainable metrics, whereby a neural network model is used to score the conversational model (typically also another neural network), see e.g. \citet{lowe2017towards,ghandeharioun2019approximating}. Such systems provide a promise of improved speed of research and development of dialogue agents, but so far have not been met with wide adoption. 
Some issues are that they  may not generalize as well to data beyond that which they are trained (overfit) and also may be biased and gameable  \cite{wu2019response,albrecht2007re}.  For a comprehensive comparison of automatic metrics -- both standard and learned metrics -- see \citet{yeh2021comprehensive}.  In general, creating a reliable automatic metric 
is still considered an open problem \cite{deriu2021survey}.

\paragraph{Crowdworkers versus experts versus organic users}
While utilizing human evaluations in research is the current standard, we contend that choosing exactly which kind of human evaluation is also still an open question. In this work we will concentrate on the study of crowdworker human evaluations, however there are several alternative paradigms. Utilizing trained experts, such as a group of researchers in the same institution, is one alternative \cite{deriu2021survey}. Compared to employing crowdworkers, while  model comparison results can agree between the two types of annotators, there can be vastly different sensitivity and win rates \cite{welleck2019neuraltext}, with the experts having more agreement and higher resulting sensitivity. On the other hand, it is harder to recruit and employ experts, limiting reproducibility. 
In both the crowdworker and expert annotator case, neither of those groups are necessarily the intended target audience of a given system. If it is possible to deploy a model to people who genuinely want to talk to it (e.g., without being paid), conversations may be more natural and evaluations will be in line with genuine interests. Evaluation by deployment can be successful \cite{gabriel2020further,shuster2020deploying}, where behavioral metrics such as the amount of conversation time per user or retention rate can serve as a proxy for interestingness and engagingness metrics. Model deployment however also has its issues. First,
user desires may not necessarily be aligned with the goals of the research itself, meaning researchers may have to develop features and improvements towards the goals of the product rather than towards long-term research. Further, experiments are difficult to set up and may be difficult to reproduce by other groups.
Crowdworker tasks can be more reproducible especially when code is made available to reproduce experiments, but there are also many pitfalls when constructing the tasks, see e.g. \citet{huynh2021survey}.

\paragraph{Conversation instructions to raters}
When utilizing evaluators in a evaluator-model conversational setup, the precise instructions on how to go about the conversation will clearly have large effects. Such instructions can control the topic, e.g. ``get to know each other'' as in the Persona-Chat task \cite{zhang2018personalizing}, versus ``have a knowledgeable conversation'' in Wizard of Wikipedia \cite{dinan2018wizard}.   Instructions can also orient workers towards a more fruitful strategy for a desired dataset, for example orienting them towards open questions on sensitive topics rather than profanity to get a bot to generate unsafe utterances \citep{xu2020recipes}. The length of the conversation will also play a role in the performance of models, for example short conversations do not test the ability of models to retain knowledge in the long-term \cite{xu2021beyond}. Overall, the style of conversation has large effects (even if the topic is unchanged) for example when instructing crowdworkers to be adversarial vs.\ non-adversarial \cite{dinan2019build}, which relates to the classic Turing Test \cite{turing1950computing}. Further, particular instruction wording choices will change the quality of conversations, as they will change how well crowdworkers understand the task  \cite{huynh2021survey}.

\paragraph{Evaluation question phrasing for raters}
Besides how the conversation is carried out, one also needs to choose the precise question (or questions) being asked to crowdworkers in order for them to rate conversations.
In open-domain conversation there are a variety of qualities one could expect from a good conversationalist, and potentially one could ask about any of them individually, as well as asking for overall performance. 
For example, \cite{see2019makes} asks evaluators for ratings of 
 interestingness, making sense, fluency, avoiding repetition, listening ability and inquisitiveness as intermediate conversational aspects, and humanness and engagingness questions to measure overall quality.
 \citet{adiwardana2020towards} ask questions based on sensibleness and specificity.
 Responsibility, toxicity and bias can also be measured \cite{xu2020recipes}.  
 Even after settling on the exact question(s) to be asked, their exact phrasing also has impact on sensitivity, as shown in \citet{li2019acute}. In that work, the authors optimized the question phrasing by running evaluations with alternative phrasings, and choosing the one with the highest agreement.

\paragraph{Rating existing versus own conversations}
The standard setup is for a human to have a conversation with a model, and rate that conversation.
Some evaluation protocols deviate from this setup, and ask evaluators to rate conversations they did not participate in. One simple approach of that kind is to present model completions of a dialogue from the fixed test set of a given task, and ask for their evaluation, with hence no human taking part in the actual conversation \cite{vinyals2015neural,Li2016AModel}.
In the Acute-Eval method \cite{li2019acute} raters are asked to compare two existing conversation logs, and the authors consider both the case of human-model chat logs, and model-model (self-chat) logs, where the former are actually a different set of human conversationalists compared to the final raters. 
\citet{deriu2020spot} considers chat logs between pairs of models, again with no humans taking part in the conversations.
These techniques allow efficient reuse of existing conversational data and have some reproducibility gains:  conversations collected in previous trials and by other systems can be directly compared with
a new system, without having to recollect additional data. This can significantly reduce the resources needed by a new evaluation, and ensure that multiple papers are comparing to
prior work consistently. On the other hand, it may be harder for evaluators to rate conversations they have not been involved in. Conversations that do not even involve humans should be treated with some scepticism, as there is no human to guide conversation and hence evaluate interactive quality. Nevertheless, such approaches do appear to be useful experimentally \cite{li2019acute,roller2021recipes}.

\paragraph{Pairwise versus single-model ratings}
Conversations are often either rated individually, 
e.g. with Likert-score ratings \cite{ram2017alexaprize,venkatesh2018evaluating,zhang2018personalizing,rashkin2019empathetic,see2019goodconversation,dinan2019second,dinan2018wizard},
or pairwise by comparing models \cite{li2019acute,liang2020beyond,vinyals2015neural,Li2016AModel,lee2020evaluation}.
 Likert scoring  relies on absolute identification rather than relative discrimination, which is less reliable in humans \citep{stewart2005absolute}, leading to different biases per annotator \citep{kulikov2018importance}. It is thus often necessary to then re-evaluate existing models at the same time as a new model, as the distribution of human annotators can easily shift over time, causing measurement errors \citep{see2019goodconversation}. Another common difficulty is related to sequential effects \cite{stewart2005absolute}, where the annotator can be influenced by the first model they evaluate, causing difficulties in using an absolute scale.
 Pairwise comparisons, on the other hand, make comparing a set of models less efficient, and also have the same problem that existing baseline models have to be essentially reassessed with respect to new ones.

\paragraph{Per-turn versus per-dialogue evaluation}
Some research  evaluates {\em single-turn} responses in conversations given gold dialogue contexts, without taking into account whole interactive conversations \cite{lee2020evaluation,vinyals2015neural,Li2016AModel}. This fails to take into account  multi-turn aspects of a conversation, for example a model repeating itself over multiple turns. {\em Per-turn} evaluation instead  conducts an entire conversation, but raters are still asked to evaluate each turn (response by their partner) \cite{adiwardana2020towards,komeili2021internet}.
Collecting per-turn evaluation also allows for measuring learning effects where workers become more adept at interacting with the bot for certain specific tasks (e.g., see \citet{xu2020recipes}).
In contrast, methods like multi-turn Likert or Acute-Eval ask evaluators to assess the entire dialogue as a whole, rather than the individual turns, under the assumption that the quality of a conversation is not simply the sum of its parts.  Literature from psychology predicts several effects when considering how people combine their impressions from single conversational turns into an evaluation of an entire conversation. The {\em primacy effect} refers to how overall judgment is more shaped by characteristics presented earlier \citep{asch1946forming,anderson1965primacy}. Conversely, the {\em recency effect} appears when more weight is given to information presented the most recently, and both effects combine to give more weight to items at the beginning and end of a list \citep{murdock1962serial,postman1965short}, with the recency effect being more prominent when judgment is elicited without any delay when the recent information is still fresh \citep{miller1959recency,hoch1984availability}.

\section{Methods}

\subsection{Evaluation techniques}

We investigate several human evaluation techniques, spanning a cross-section of the different methods discussed in existing work. Specifically:
\begin{itemize}
    \item Single-model per-turn evaluations
    \item Single-model per-dialogue evaluations
    \item Pairwise per-turn evaluations
    \item Pairwise per-dialogue evaluations
    \item Pairwise per-dialogue {\em self-chat} evaluations
\end{itemize}

\begin{figure*}[t!]
\center
\includegraphics[width=\textwidth]{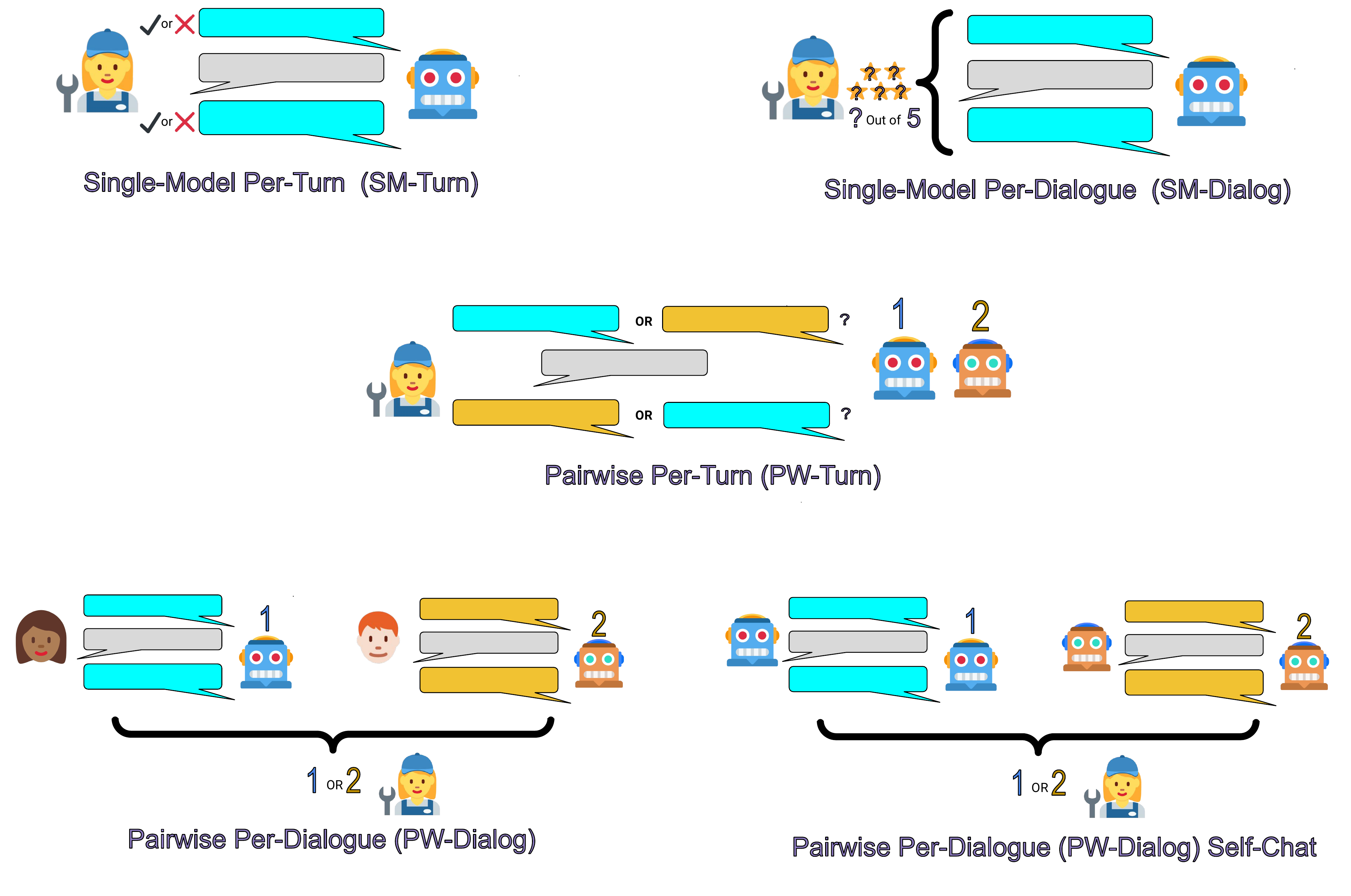}
  \caption{The human evaluation methods we compare in this work. \smturn{} rates each bot response during the conversation, while \smdialog{} rates the entire conversation. \pwturn{} compares two different bots' responses at every turn in the conversation, while \pwdialog{} compares two entire conversations with two different bots. \pwdialog{} self-chat compares two conversations which only involve the two bots talking to themselves (self-chat). }
  \label{image:model_figs}
\end{figure*}

We thus compare the spectrum of single vs.\ pairwise and per-turn vs.\ per-dialogue variations,  as well as trying a self-chat method compared to conventional human-bot conversation ratings. Figure \ref{image:model_figs} summarizes the methods.
In the following, we will describe our exact methodology for each.

\subsubsection{Conversational setting}

Our human-bot evaluations consist of a set of conversations. Each conversation is between a crowdworker 
(the ``Human Speaker'') paired with a conversational model (the ``Bot Speaker''). The Human Speaker will speak naturally in the conversation, and they will be role-playing as a certain persona with the help of two provided {\em persona sentences} given to them at the start of the conversation, see Figure~\ref{image:crowdsourcing} (left) for an example.

The Human Speaker's first message in the conversation is fixed to \textit{``Hi!''}, following the convention of \citet{adiwardana2020towards}. The conversation ends after the Human Speaker and Bot Speaker have both spoken for 6 turns each. 
We test three different evaluation metrics, {\em preference}, {\em humanness} and {\em interestingness}, with exact wordings described in the following subsections.

\subsubsection{Pairwise per-turn evaluations}
\label{sec:methods_pwpt}

\begin{figure*}[t!]
\center
\includegraphics[width=\textwidth]{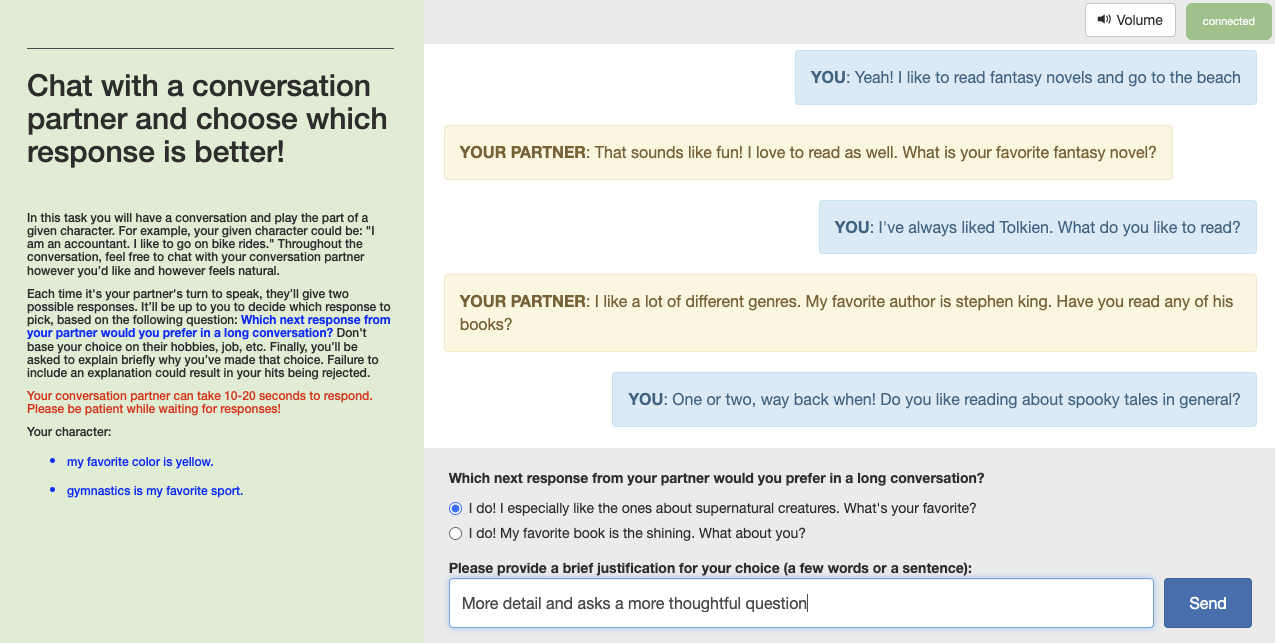}
  \caption{Screenshot of the Pairwise Per-Turn (\pwturn{}) evaluation technique, in which we ask crowdworkers to choose one of two possible responses from their conversation partner and describe why that response is better. The two responses come from the two models that we are trying to compare the performance of.}
  \label{image:crowdsourcing}
\end{figure*}

The Pairwise Per-Turn evaluation (\pwturn{}) technique provides annotations for every turn of conversation by asking for the crowdworker to choose from a pair of model responses after every sent message.
Hence, in this setting the Human Speaker speaks to a Bot Speaker, the latter of which  actually represents the two models to be compared. The Human Speaker will speak naturally in the conversation. Every time that it is the Bot Speaker's turn to speak, the crowdworker will first be presented with two options as possible responses: each response will come from one of the two models being compared, similarly to \citet{clark2021choose}. We randomize the ordering of these model responses. The worker must choose the better response for the given evaluation metric. The wordings we use for the three metrics are adapted from \citet{li2019acute}:

\begin{itemize}
  \item \textbf{Preference}: \textit{“Which next response from your partner would you prefer in a long conversation?”}
  \item \textbf{Humanness}: \textit{“Which next response from your partner sounds more human?”}
  \item \textbf{Interestingness}: \textit{“If you had to say one of these responses is interesting and one is boring, which would you say is more interesting?”}
\end{itemize}

The worker must give a free-text justification for their choice of response. The response that they choose is set to be the actual response given by the Bot Speaker, and the conversation continues from there.
Figure~\ref{image:crowdsourcing} provides a screenshot example of the UI.
 A description of quality checks performed when onboarding workers for this evaluation technique is given in  Appendix~\ref{sec:onboarding}.
In our experiments we will consider win rates based on simply averaging over turns, as well as nonlinear combinations of per-turn results over entire dialogues (e.g., winner-takes-all voting) in order to measure their impact.

\subsubsection{Pairwise per-dialogue evaluations}

\begin{figure*}[t!]
\center
\includegraphics[width=\textwidth]{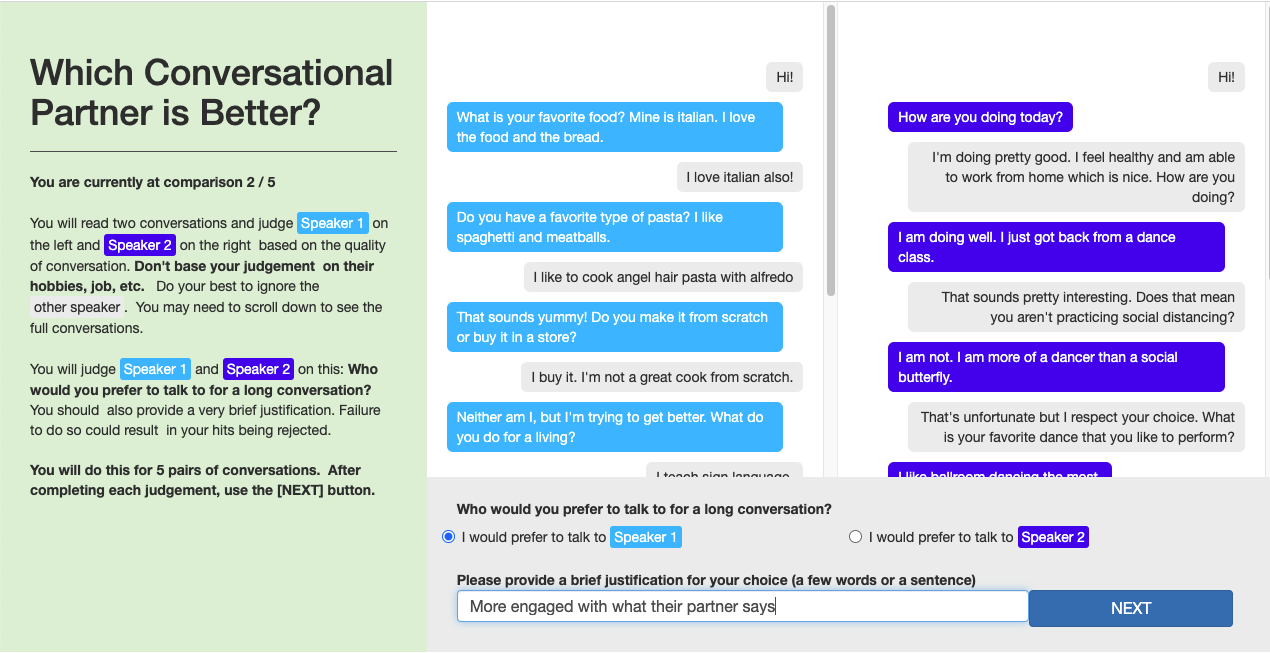}
  \caption{Screenshot of the Pairwise Per-Dialogue (\pwdialog{}) evaluation technique, in which we ask crowdworkers to choose which of two speakers in two separate conversations is better on the given evaluation metric, here \textit{``Who would you prefer to talk to for a long conversation?''} The crowdworker must then describe why that speaker is better.}
  \label{image:acute_screenshot}
\end{figure*}

The Pairwise Per-Dialogue evaluation (\pwdialog{}) technique we introduce asks evaluators to choose between two models by presenting a pair of conversations. The method we employ is identical to the Acute-Eval method \citep{li2019acute}, but for consistency with the names of other techniques, we refer to them here as \pwdialog{} evaluations.
For each of the model pairs and evaluation metrics used, we collect evaluations on (1) conversations conducted between a crowdworker and a model agent; and (2) self-chat conversations conducted between two conversational agents of the same model (the \textit{self-chat} variant).
The wordings we use (from \citep{li2019acute}) are almost identical to the \pwturn{} versions, but phrased for the per-dialogue, rather than per-turn, case:

\begin{itemize}
  \item \textbf{Preference}: \textit{“Who would you prefer to talk to for a long conversation?”}
  \item \textbf{Humanness}: \textit{“Which speaker sounds more human?”}
  \item \textbf{Interestingness}: \textit{“If you had to say one of these speakers is interesting and one is boring, who would you say is more interesting?”}
\end{itemize}

Figure~\ref{image:acute_screenshot} provides a screenshot example of the UI.

\subsubsection{Single-model evaluations}

\begin{figure*}[t!]
\center
\includegraphics[width=\textwidth]{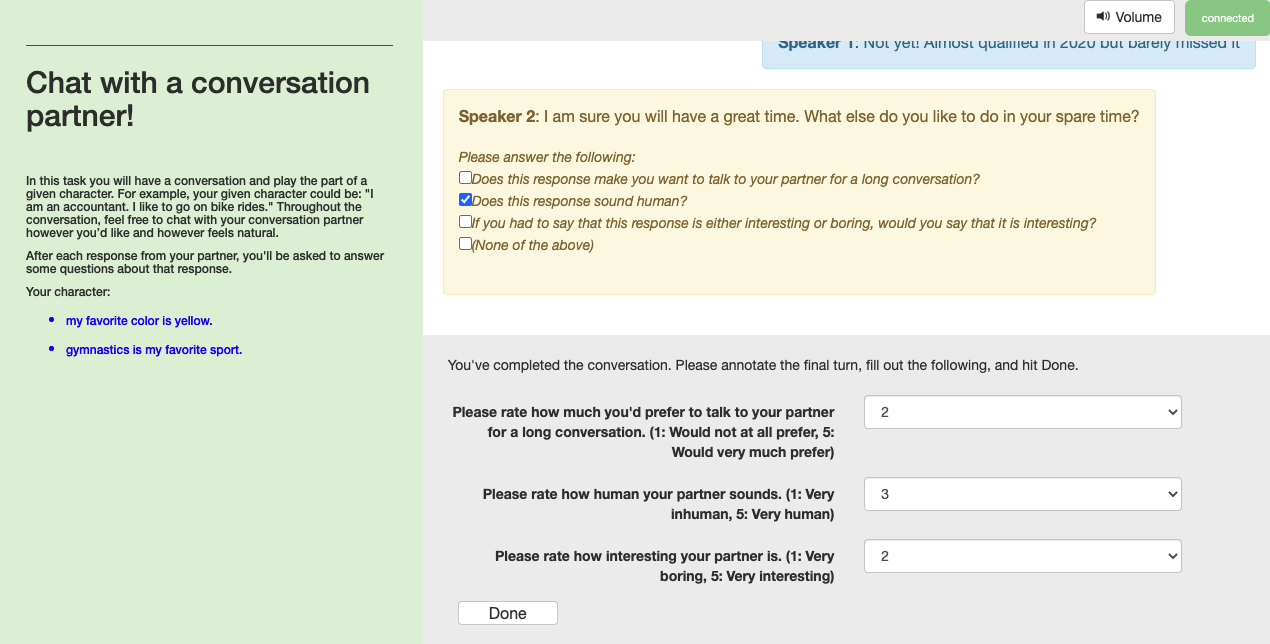}
  \caption{Screenshot of the crowdsourcing task for collecting Single-Model Per-Turn (\smturn{}) and Single-Model Per-Dialogue (\smdialog{}) evaluations. We ask the crowdworker to annotate each response from their partner along several dimensions, as well as give a global Likert-scale rating of their partner's performance at the end of the conversation.}
  \label{image:per_turn_single_model}
\end{figure*}

In our single-model evaluation experiments, we combine per-turn and per-dialogue into the same UI (see Figure~\ref{image:per_turn_single_model} for a screenshot).\footnote{This may have undesirable effects in correlating their results, but nonetheless they do appear to perform quite differently in evaluations.}
A crowdworker chats with a conversational agent backed by a single model, and for each response of that model the worker must annotate whether it is engaging, human-like, and/or interesting, with wording provided in the screenshot. At the end of the conversation, again consisting of 6 messages per speaker, the worker must rate their partner on a Likert scale of 1 to 5 for each of the three evaluation metrics listed in Section~\ref{sec:methods_pwpt}. We refer to the per-turn annotations of model responses from this task as Single-Model Per-Turn evaluations (\smturn{}) and the end-of-conversation Likert scores as Single-Model Per-Dialogue evaluations (\smdialog{}).


Empirically, we find that \smturn{} success rates and \smdialog{} Likert scores are highly dependent on the particular day that the evaluations are collected: this is perhaps due to day-to-day variability in the pool of crowdworkers. To counteract this, we run these evaluations on all four of the models discussed in this work (Section~\ref{sec:models}) simultaneously.\footnote{For the pairwise evaluation techniques \pwturn{} and \pwdialog{}, we collect evaluations over several days across multiple weeks for each of the three model pairs evaluated. This helps to smooth out variability among days.}

\subsection{Quality checks on crowdworkers}
\label{sec:quality_checks}

In order to ensure that our comparisons between evaluation techniques are not affected by variability in the pool of crowdworkers when running one technique vs.\ another, we adopt a consistent set of criteria across all techniques regarding which workers to exclude from our final set of data. If a worker fails one of the checks in Appendix~\ref{sec:filtering_checks} during one of the per-turn evaluations \pwturn{} or \smturn{}, we retroactively exclude their ratings from all of the evaluation techniques.

In order to prevent any worker from disproportionately contributing to the final results, each worker is restricted to one conversation per model pair and evaluation metric (for \pwturn{} and \pwdialog{}) or one conversation per model (for \smturn{} and \smdialog{}). All evaluations are collected among residents of the United States on weekdays, from roughly 9 AM to 6 PM in the U.S. Eastern time zone, following \citet{li2019acute}.

\subsection{Models}
\label{sec:models}

We analyze the relative performance of these five human evaluation techniques, \smturn{}, and \smdialog{}, \pwturn{}, \pwdialog{} and \pwdialog{} {\em self-chat},  on four different well-performing but relatively similar dialogue models from \citet{roller2021recipes}:
\begin{itemize}
  \item \textbf{BlenderBot3B}: The version of BlenderBot with 2.7 billion parameters, pretrained on a previously existing Reddit dataset extracted and obtained by a third party and made available on pushshift.io \citep{baumgartner2020pushshift} and then fine-tuned on several purpose-built dialogue datasets.
   \item \textbf{BlenderBot3B-M0}: BlenderBot3B uses a minimum generation length of 20 tokens to ensure relatively long, interesting responses.
  We also compare to  exactly the same model but  without a minimum generation length, referring to it with  \textbf{-M0} postfix.
  \item \textbf{BlenderBot90M}: The variant of BlenderBot with 90 million parameters, trained on the same datasets as BlenderBot3B.
  \item \textbf{Reddit3B}: BlenderBot3B, but only pretrained on the third-party Reddit dump and not fine-tuned on dialogue datasets.
\end{itemize}

For all models, we use the same generation settings as in \citet{roller2021recipes}, apart from the \textbf{-M0}  adaptation. 
We choose these relatively similar models in our experiments as a difficult challenge for evaluation techniques to tell which one is best.

For the two pairwise evaluation techniques, we specifically perform comparisons between three pairs of models, each of which differ in a characteristic way:
\begin{itemize}
  \item \textbf{Length comparison}: Comparing BlenderBot3B to BlenderBot3B-M0: these models differ only in the length of their generations.
  \item \textbf{Size comparison}: Comparing two models with different numbers of parameters, BlenderBot3B and BlenderBot90M.
  \item \textbf{Fine-tuning comparison}: Comparing the fine-tuned BlenderBot3B to the pretrained-only Reddit3B (both with the same number of parameters).
\end{itemize}

\section{Results}

\subsection{Evaluation data collection}
\label{sec:data_collection}

After filtering out workers with unacceptable messages following Section \ref{sec:quality_checks}, we are left with a minimum of 144 and a mean of 231 ratings (typically 6 per conversation) for each of the \pwturn{} evaluations, a minimum of 191 and a mean of 324 ratings for \pwdialog{}, a minimum of 349 and a mean of 411 ratings (typically 6 per conversation) for \smturn{}, and a minimum of 58 and a mean of 68 ratings for \smdialog{} evaluations (for which there is only one rating per conversation). On average, the collection of ratings after filtering represents 5.73 hours of worker labor for \pwturn{} per model pair and evaluation metric, 6.03 hours for \pwdialog{} per model pair and evaluation metric, and 4.39 hours for joint \smturn{}/\smdialog{} evaluations per model.

\subsection{Model win rates from pairwise per-turn evaluations}
\label{sec:win_rates}

\begin{table}[h!]
\centering
\begin{small}
\begin{tabular}{cccccc}
\toprule
& & All turns & \multicolumn{3}{c}{Turns 2 to 6} \\
\cmidrule(lr){3-3} \cmidrule(lr){4-6}
Comp. & Metric & Lin & Lin & Sqr & WTA \\
\midrule
Length & Pref & 63\% & 67\% & 72\% & \textbf{74\%} \\
 & Human & 63\% & 68\% & 75\% & \textbf{79\%} \\
 & Inter & 68\% & 70\% & 77\% & \textbf{84\%} \\
\midrule
Size & Pref & 48\% & 52\% & \textbf{53\%} & 49\% \\
 & Human & 51\% & 56\% & \textbf{58\%} & 54\% \\
 & Inter & 49\% & 52\% & 54\% & \textbf{55\%} \\
\midrule
FT & Pref & 80\% & 82\% & 88\% & \textbf{93\%} \\
 & Human & 81\% & 84\% & 88\% & \textbf{93\%} \\
 & Inter & 71\% & 75\% & 80\% & \textbf{85\%} \\
\bottomrule
\end{tabular}
\end{small}
\caption{\pwturn{} win rates of BlenderBot3B vs.\ BlenderBot3B-M0 (``Length''), vs.\ BlenderBot90M (``Size''), and vs.\ the base pretrained model, Reddit3B (``FT''), across three different evaluation metrics, Preference, Humanness, and Interestingness. Win rates are computed both across all turns and across only the last 5 turns from the Bot Speaker (``Turns 2 to 6''). \textbf{Lin}: the linear win rate $x/(x+y)$ of BlenderBot3B, given $x$ wins of BlenderBot3B and $y$ wins of the comparison model. \textbf{Sqr}: the ``squared'' win rate $x^2/(x^2+y^2)$, calculated per-conversation and then averaged across all conversations. \textbf{WTA}: the winner-takes-all win rate, defined as the percentage of all conversations for which BlenderBot3B wins on more turns, or equivalently $x^\infty{}/(x^\infty{}+y^\infty{})$ as calculated per-conversation. Winner-takes-all scores are generally highest (highest values bolded).}
\label{tab:win_rates_pwturn}
\end{table}

We compute the win rates of BlenderBot3B over other models in Table~\ref{tab:win_rates_pwturn} for the pairwise evaluation technique \pwturn{}. We expect BlenderBot3B to be better, hence values closer to 100\% are deemed more preferable. We  display the win rates of four different variants: including all 6 conversation turns from the Bot Speaker, excluding the Bot Speaker's first turn from the evaluations, and computing a nonlinear function of the turns: either calculating squared or winner-takes-all win rates for each conversation and then averaging those scores across all conversations.
We generally find that \pwturn{} win rates are higher when dropping the first turn of the Bot Speaker, as discussed further in Section \ref{sec:win_rates_per_turn}. Win rates are typically even higher by aggregating over conversations in a winner-takes-all fashion, which has the effect of reducing the turn-by-turn variability of which model's response is chosen by the crowdworker.

We find that, in general, win rates of BlenderBot3B do not vary much as a function of the evaluation question  used when asking workers to choose one model response over the other. It is unclear \textit{a priori} whether this results from an ambiguity in the precise definitions of these questions/metrics when interpreted by the workers, correlations in how well models perform on some metrics vs.\ others, or some other reason.



\subsubsection{Model win rates as a function of turn}
\label{sec:win_rates_per_turn}

\begin{figure}[h!]
\center
\includegraphics[width=\columnwidth]{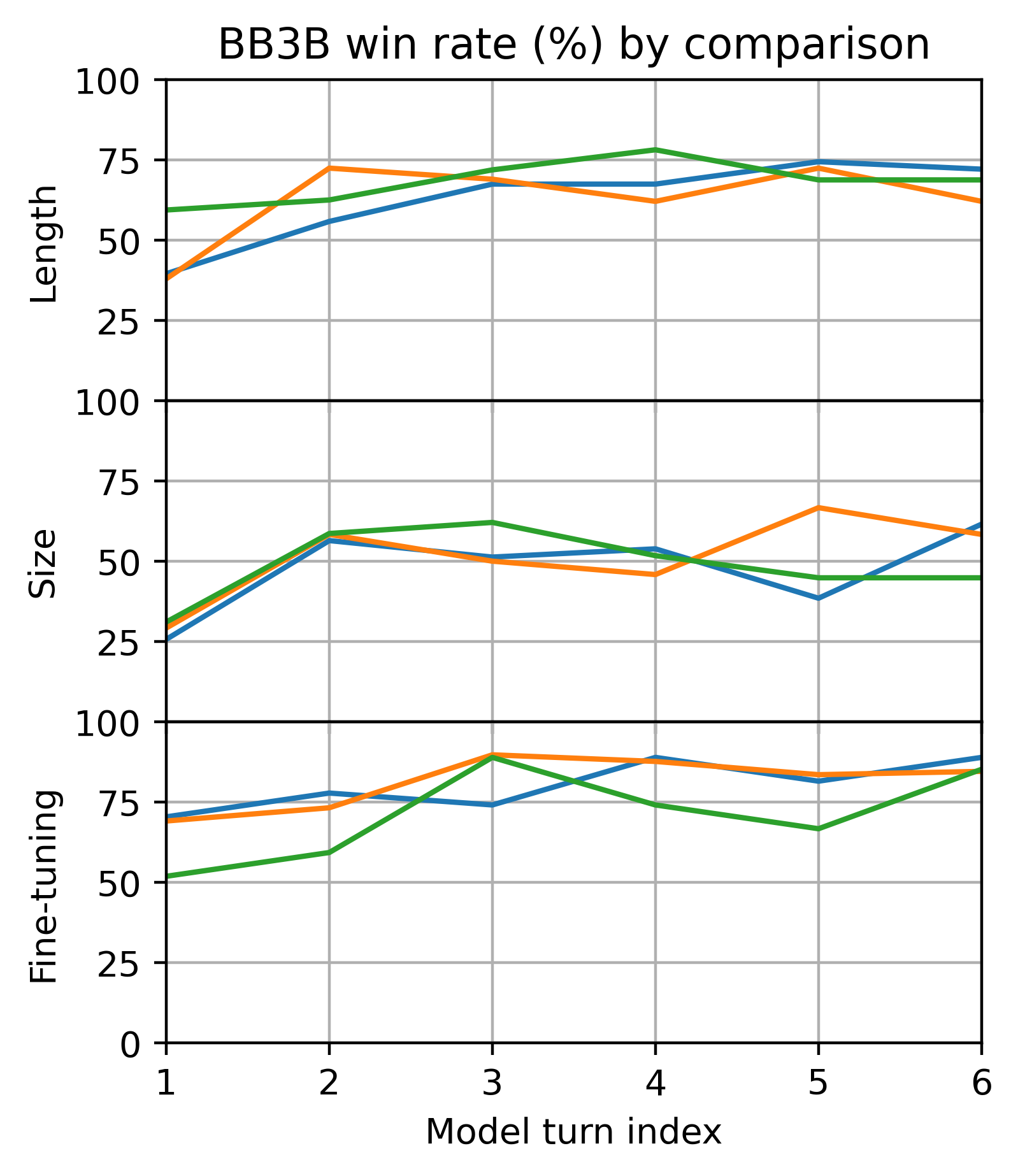}
  \caption{Win rate of BlenderBot3B vs.\ other models for the comparisons in Section \ref{sec:models} on the \pwturn{} evaluations, as a function of the number of Bot Speaker turns into the conversation, for the Preference (blue), Humanness (orange), and Interestingness (green) metrics. BlenderBot3B tends to fare better against other models in later turns of the conversation.}
  \label{image:win_rate_by_turn_idx}
\end{figure}

Unlike \pwdialog{}, the \pwturn{} technique is able to measure differences in the win rate of models as a function of the number of turns into the conversation. In Figure~\ref{image:win_rate_by_turn_idx} and Table~\ref{tab:win_rate_by_turn_idx}, we see that BlenderBot3B's win rates tend to be closer to 50\% in the first 1 or 2 turns of the Bot Speaker, and higher later: this may be because the first few lines of the conversation typically consist of greetings (\textit{``Hi, how are you?''}) or pleasantries, which may be harder to judge model performance on. 
However, it may also be because improvements are accumulated and factored into evaluators' decisions later in the conversation.
Strikingly, BlenderBot3B performs very poorly vs.\ BlenderBot90M (the Size comparison) on the first Bot Speaker turn: empirically, this may be due to the fact that BlenderBot3B generally starts its first responses with the greetings \textit{``Hi''} or \textit{``Hello''} much less frequently than BlenderBot90M does.



\subsection{Model scores from single-model evaluations}
\label{sec:single_model_results}

\begin{table}[h!]
\centering
\begin{small}
\begin{tabular}{cccccc}
\toprule
& & \multicolumn{3}{c}{\smturn{}} & \smdialog{} \\
\cmidrule(lr){3-5} \cmidrule(lr){6-6}
& & All & \multicolumn{2}{c}{Turns 3 to 6} & \\
\cmidrule(lr){3-3} \cmidrule(lr){4-5}
Met. & Model & Lin & Lin & WTA & \\
\midrule
\multirow{4}{*}{\rotatebox[origin=c]{90}{Preference}} & BB3B & 70\% & \textbf{71\%} & \textbf{73\%} & \textbf{4.19} \\
 & BB3B-M0 & \textbf{71\%} & 70\% & 70\% & 4.02 \\
 & BB90M & 65\% & 64\% & 65\% & 3.97 \\
 & Reddit3B & 55\% & 50\% & 50\% & 3.30 \\
\midrule
\multirow{4}{*}{\rotatebox[origin=c]{90}{Human}} & BB3B & \textbf{70\%} & \textbf{72\%} & \textbf{73\%} & \textbf{4.49} \\
 & BB3B-M0 & 67\% & 66\% & 70\% & 4.22 \\
 & BB90M & 65\% & 66\% & 70\% & 3.94 \\
 & Reddit3B & 56\% & 54\% & 53\% & 3.50 \\
\midrule
\multirow{4}{*}{\rotatebox[origin=c]{90}{Interesting}} & BB3B & \textbf{44\%} & \textbf{45\%} & \textbf{47\%} & \textbf{4.22} \\
 & BB3B-M0 & 35\% & 35\% & 36\% & 3.76 \\
 & BB90M & 39\% & 40\% & 42\% & 3.83 \\
 & Reddit3B & 39\% & 39\% & 37\% & 3.30 \\
\bottomrule 
\end{tabular}
\end{small}
\caption{Performance of BlenderBot3B (BB3B), BlenderBot3B-M0 (BB3B-M0), BlenderBot90M (BB90M), and Reddit3B on \smturn{} and \smdialog{} evaluations. \smturn{} mean success rates are calculated across all turns (``All'') or across only the last 4 turns from the Bot Speaker (``Turns 3 to 6''). Scores represent the overall fraction of model responses marked as successful on the given evaluation metric (``Lin'') or the number of conversations for which at least half of the model responses are marked as successful (winner-takes-all, ``WTA''). \smdialog{} evaluations are Likert scores (with standard deviations in the range of 0.8 to 1.3). Highest scores across models are bolded.}
\label{tab:win_rate_likert_yes_no}
\end{table}

Table~\ref{tab:win_rate_likert_yes_no} provides the per-turn success rates (\smturn{}) and end-of-conversation Likert scores (\smdialog{}) over all models for various metrics. As with the pairwise evaluations of Section~\ref{sec:win_rates} and \citet{roller2021recipes}, BlenderBot3B generally outperforms the other models using the \smturn{} and \smdialog{} methods as well. 
Table~\ref{tab:smturn_success_rate_by_turn_idx} (in the Appendix) shows success rates from the \smturn{} technique as a function of conversation turn (rather than aggregated). BlenderBot3B scores are generally stable across conversation turn but are slightly lower on the first two turns of the Bot Speaker, echoing similar findings with \pwturn{} in Section~\ref{sec:win_rates_per_turn}. We thus also consider removing \smturn{} scores from the first two turns in order to maximize the performance of BlenderBot3B relative to the other models. As with \pwturn{}, we find that calculating the winner-takes-all score per conversation allows for an even bigger separation in performance between BlenderBot3B and the other models. 

 Unlike \pwturn{} for which win rates are similar across all three evaluation metrics (Section~\ref{sec:win_rates}), single-model success rates on the Interestingness metric are generally lower than those on the other two, especially for \smturn{}. We hypothesize that the juxtaposition of all three evaluation questions side-by-side in the UI of the \smturn{} and \smdialog{} crowdworker task (Figure \ref{image:per_turn_single_model}) may aid workers in distinguishing among these three metrics and rating models differently on them. 



\subsubsection{Relationship between per-turn ratings and final ratings}

\begin{figure}[h!]
\center
\includegraphics[width=\columnwidth]{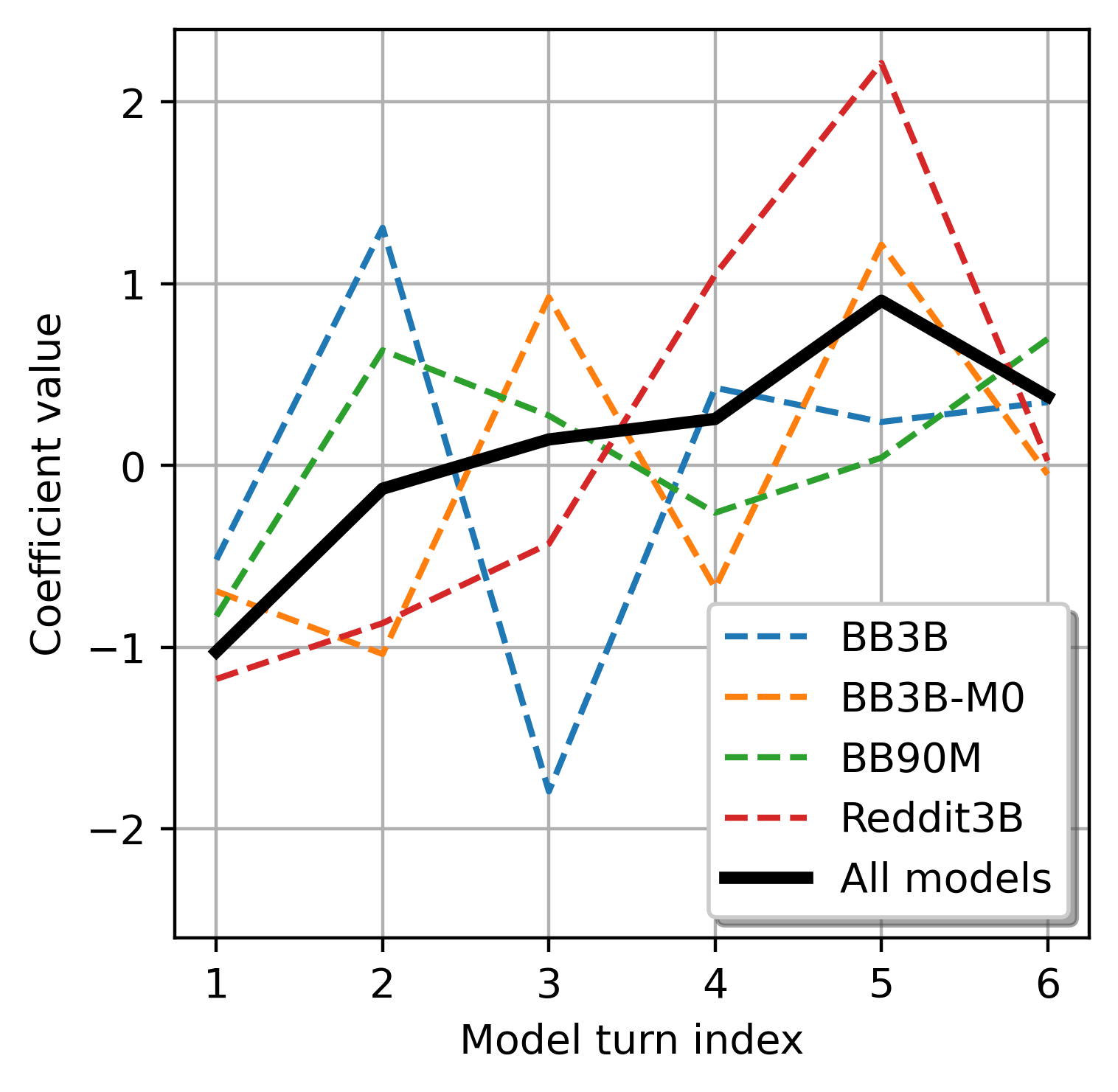}
  \caption{The per-turn coefficients of \smturn{} success rates in an OLS regression with \smdialog{} Likert scores as the dependent variable. \smturn{} rates and \smdialog{} scores are averaged across evaluation metrics. The black curve represents data from all models concatenated together. \smturn{} rates from later turns tend to be more positively correlated to the final \smdialog{} Likert scores, suggesting a possible recency bias.}
  \label{image:likert_regression}
\end{figure}

Given that \smturn{} allows us to measure per-turn ratings of model performance, it is worth exploring whether there are certain turns of the conversation that contribute more strongly to the workers' final Likert-scale ratings of the conversation (\smdialog{}). Figure~\ref{image:likert_regression} plots the coefficients of workers' per-turn \smturn{} ratings in an OLS regression, with the \smdialog{} Likert score at the end of the conversation as the dependent variable. (Here, we reduce variability by taking the mean over the three evaluation metrics for each turn's \smturn{} ratings and \smdialog{} Likert scores.) Generally, we see a higher positive coefficient of the \smturn{} ratings in later turns in the conversation, which implies that the workers may have a recency bias: they may remember the most recent turns of the conversation more strongly when determining how to rate the model's performance overall.

\subsection{Direct comparison of all evaluation techniques}
\label{sec:comparing_all_techniques}

In this section we directly compare all the pairwise and single-model evaluation techniques to each other to discern their relative strengths.

\subsubsection{Computing win rates across all techniques}
\label{sec:win_rates_all_techniques}

\begin{table*}[h!]
\centering
\begin{small}
\begin{tabular}{cccccccc}
\toprule
& & \pwturn{} & \multicolumn{2}{c}{\pwdialog{}} & PW combo & \smturn{} & \smdialog{} \\
\cmidrule(lr){3-3} \cmidrule(lr){4-5} \cmidrule(lr){6-6} \cmidrule(lr){7-7} \cmidrule(lr){8-8}
Comparison & Metric & Turns 2--6, WTA & Human & Self & & Turns 3--6, WTA & \\
\midrule
Length & Pref & 74\% & 77\% & \textbf{82\%} & 80\% & 55\% & 58\% \\
 & Human & 79\% & 77\% & \textbf{83\%} & 81\% & 52\% & 59\% \\
 & Inter & 84\% & \textbf{85\%} & 73\% & 73\% & 60\% & 65\% \\
\midrule
Size & Pref & 49\% & 56\% & 55\% & 54\% & 59\% & \textbf{60\%} \\
 & Human & 54\% & 61\% & 55\% & 55\% & 52\% & \textbf{66\%} \\
 & Inter & 55\% & 59\% & 57\% & 56\% & 55\% & \textbf{64\%} \\
\midrule
Fine-tuning & Pref & \textbf{93\%} & 70\% & 66\% & 69\% & 64\% & 71\% \\
 & Human & \textbf{93\%} & 54\% & 61\% & 65\% & 62\% & 73\% \\
 & Inter & \textbf{85\%} & 59\% & 64\% & 66\% & 60\% & 70\% \\
\bottomrule
\end{tabular}
\end{small}
\caption{Win rates of BlenderBot3B vs.\ other models, for all evaluation techniques. For the per-turn techniques \pwturn{} and \smturn{}, only the specified Bot Speaker turns are used to compute winner-takes-all scores, as in Tables~\ref{tab:win_rates_pwturn} and~\ref{tab:win_rate_likert_yes_no}. We show \pwdialog{} win rates as measured on conversations between a crowdworker and a model (``Human'') as well as from model self-chats (``Self''). ``PW combo'' represents the win rate when sampling ratings from \pwturn{} (turns 2--6) and \pwdialog{} (on model self-chats) at a ratio of 1:5. \pwturn{}, \pwdialog{}, and \smdialog{} are each found to be most sensitive at measuring model performance for one of the three model comparisons tested (highest win rates bolded). See Section~\ref{sec:data_collection} for the number of evaluations and the estimated total number of worker-hours per technique.}
\label{tab:win_rates_compare_techniques}
\end{table*}

In order to directly compare the performance of \smturn{} and \smdialog{} with that of the pairwise techniques, we calculate effective win rates for the two single-model techniques by bootstrapping samples of ratings from different models and then calculating how often \smturn{} success rates and \smdialog{} Likert scores from one model are higher than those of another. Following the analysis of best performing methods from Sections~\ref{sec:win_rates} and~\ref{sec:single_model_results}, we consider only Bot Speaker turns 2 through 6 for \pwturn{} and turns 3 through 6 for \smturn{} in winner-takes-all (WTA) mode, in order to maximize the ability of these techniques to distinguish different models' performances.

Table~\ref{tab:win_rates_compare_techniques} compares the win rates produced by all evaluation techniques. Overall, we find that a different technique performs best for each of the three model comparison types:
\begin{itemize}
  \item \textbf{Length comparison}: The pairwise evaluation techniques \pwdialog{} and \pwturn{} perform much better than the single-model ones. BlenderBot3B responses tend to contain many more words on average than those of BlenderBot3B-M0, and so we hypothesize that this difference in sensitivity among the techniques may be due to the fact that viewing responses from both models side-by-side makes the length differences between them much more evident, especially when comparing two entire conversations as in \pwdialog{}. Thus, if crowdworkers tend to prefer longer responses on average, the side-by-side comparison of model responses might aid in their ability to choose BlenderBot3B responses over those of BlenderBot3B-M0.
  
  \item \textbf{Size comparison}: The differences among the techniques here are smaller than for the Length comparison, with the full-dialogue techniques \pwdialog{} and \smdialog{} slightly outperforming the per-turn ones. As shown by \citet{roller2021recipes}, BlenderBot3B and BlenderBot90M do not perform statistically significantly differently on Acute-Evals (i.e. \pwdialog{}) on self-chat conversations. Thus, it may make sense that any small differences in performance between these models are more evident on the level of whole conversations.
  
  \item \textbf{Fine-tuning comparison}: In this comparison, \pwturn{} performs best out of all techniques. Because the Reddit3B model was not fine-tuned on conversational dialogue datasets, its responses to its partner generally make less sense in context than those of BlenderBot3B. We hypothesize that these more nonsensical responses may be very obvious to workers who are in the middle of having a conversation with the Bot Speaker during the \pwturn{} evaluation. However, these responses may be less obvious to workers reading whole conversations in the \pwdialog{} evaluation who have not interacted with the models directly, as well as to workers in \smturn{} and \smdialog{} evaluations who cannot directly compare Reddit3B responses to those of a model that has been fine-tuned on dialogue.
  
\end{itemize}

\paragraph{Explainability in experiments: analysis of crowdworker reasons}

During the crowdworker evaluation tasks, we also ask for reasons for the crowdworker's judgments. These reasons can give interpretability to the results.
A full analysis is given in Appendix \ref{sec:justification}.
Overall, we find justifications that make sense in each of the three model comparisons, e.g.\ in the length comparison we see keywords like {\em ``information''} and {\em ``detailed''} appearing often. For the fine-tuning comparison, we often find keywords like {\em ``flows''}, {\em ``personal''} and {\em ``contradicts''}, which shows that the fine-tuning conversational datasets like Persona-Chat provide for more personal, less contradictory, and flowing conversations.

\paragraph{Repeatability of experiments}

We provide an analysis in Appendix~\ref{sec:variability} of the variability of model win rates over time for each of the evaluation techniques. Overall, we find that 
\pwturn{}, \pwdialog{}, and \smturn{} vary least across chunked experiments, with \smdialog{} having more variability. This makes the use of \smdialog{} less compelling.

\subsubsection{Overall findings}
The results of these three model comparisons hint that perhaps a per-turn evaluation technique may be more suitable for pairs of models that differ in their ability to reply sensibly in a way that is easily detectable by their partner (i.e. BlenderBot3B vs.\ Reddit3B), but that a whole-conversation technique may be preferable when differences between models are more sensitive. However, evaluations on many more pairs of models would be needed to sufficiently support such a broad hypothesis. We also find that single-model techniques perform competitively to pairwise ones except for when model generations differ by average length: in this case, comparing the responses of both models side-by-side may make the differences between them more apparent than just viewing them separately.

\paragraph{Combining techniques}

Given how much the relative sensitivities of different evaluation techniques vary across different pairs of models, we also explore whether combining results from multiple techniques together may allow for a compromise technique that performs reasonably well in all cases. We thus include in Table~\ref{tab:win_rates_compare_techniques} the win rate (``PW combo'') when sampling ratings from the \pwturn{} and \pwdialog{} techniques together at a ratio of 1:5. This sampling retains most of the ability of \pwdialog{} to quickly compare BlenderBot3B to BlenderBot3B-M0 and BlenderBot90M (the Length and Size comparisons), and it also gains some of \pwturn{}'s superior strength at measuring the performance of BlenderBot3B over Reddit3B (the Fine-tuning comparison).

By contrast, since ratings for the two single-model techniques \smturn{} and \smdialog{} are collected simultaneously, ratings from both techniques on a given conversation can be averaged together to achieve slightly finer sensitivity than either technique individually. Figures~\ref{image:time_to_sig__sm__length}, \ref{image:time_to_sig__sm__size}, and~\ref{image:time_to_sig__sm__fine_tuning} show that, with the proper weighting, such averaging can produce a statistically significant difference between models a bit faster than that of \smdialog{} and dramatically faster than that of \smturn{} (Section~\ref{sec:time_to_sig_single_model}).

\subsubsection{Crowdsourcing time needed}
\label{sec:time_to_sig}

\begin{figure}[h!]
\center
\includegraphics[width=\columnwidth]{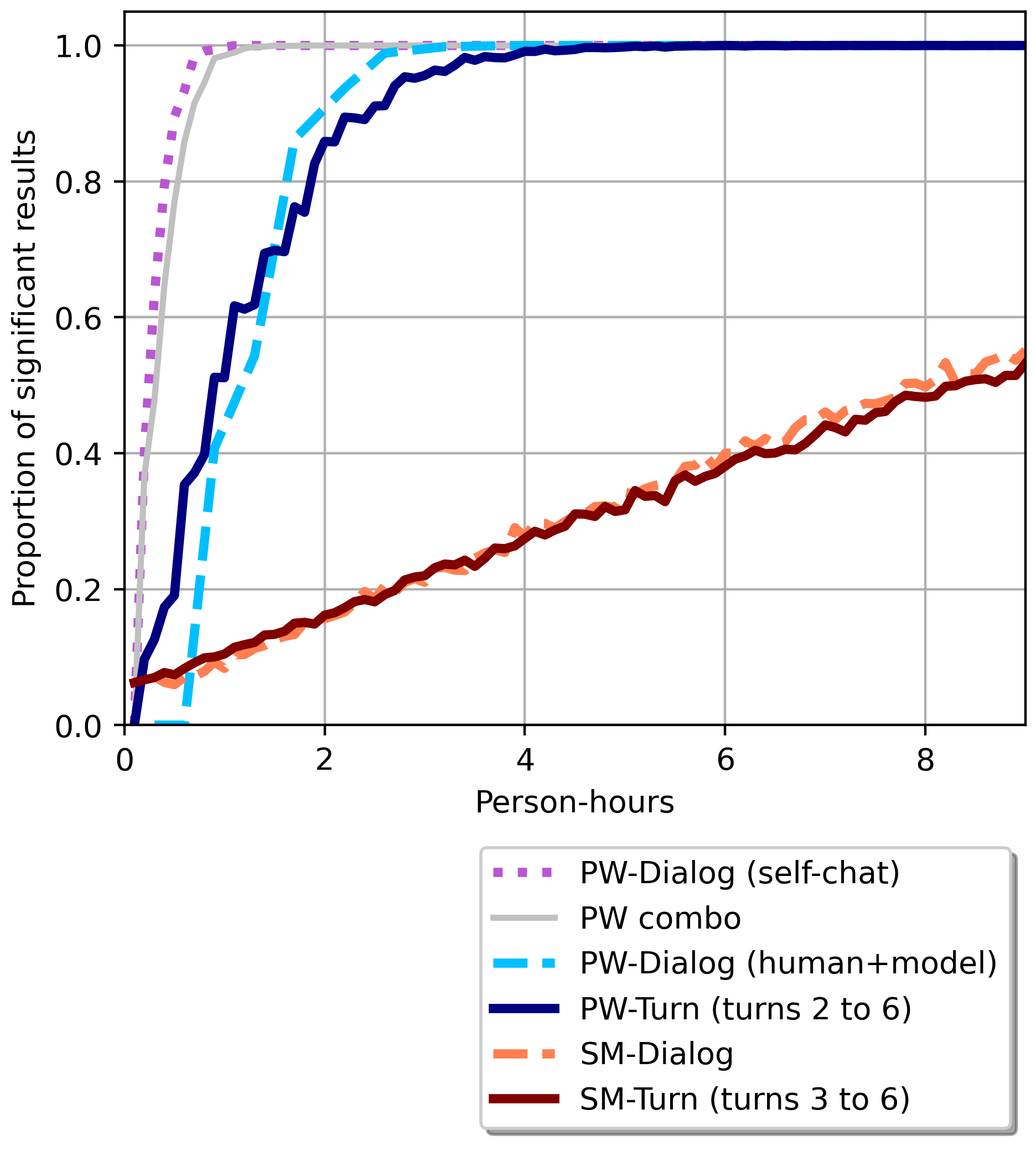}
  \caption{The time needed for statistical significance for the Length comparison between models (BlenderBot3B vs.\ BlenderBot3B-M0).}
  \label{image:time_to_significance__length}
\end{figure}

\begin{figure}[h!]
\center
\includegraphics[width=\columnwidth]{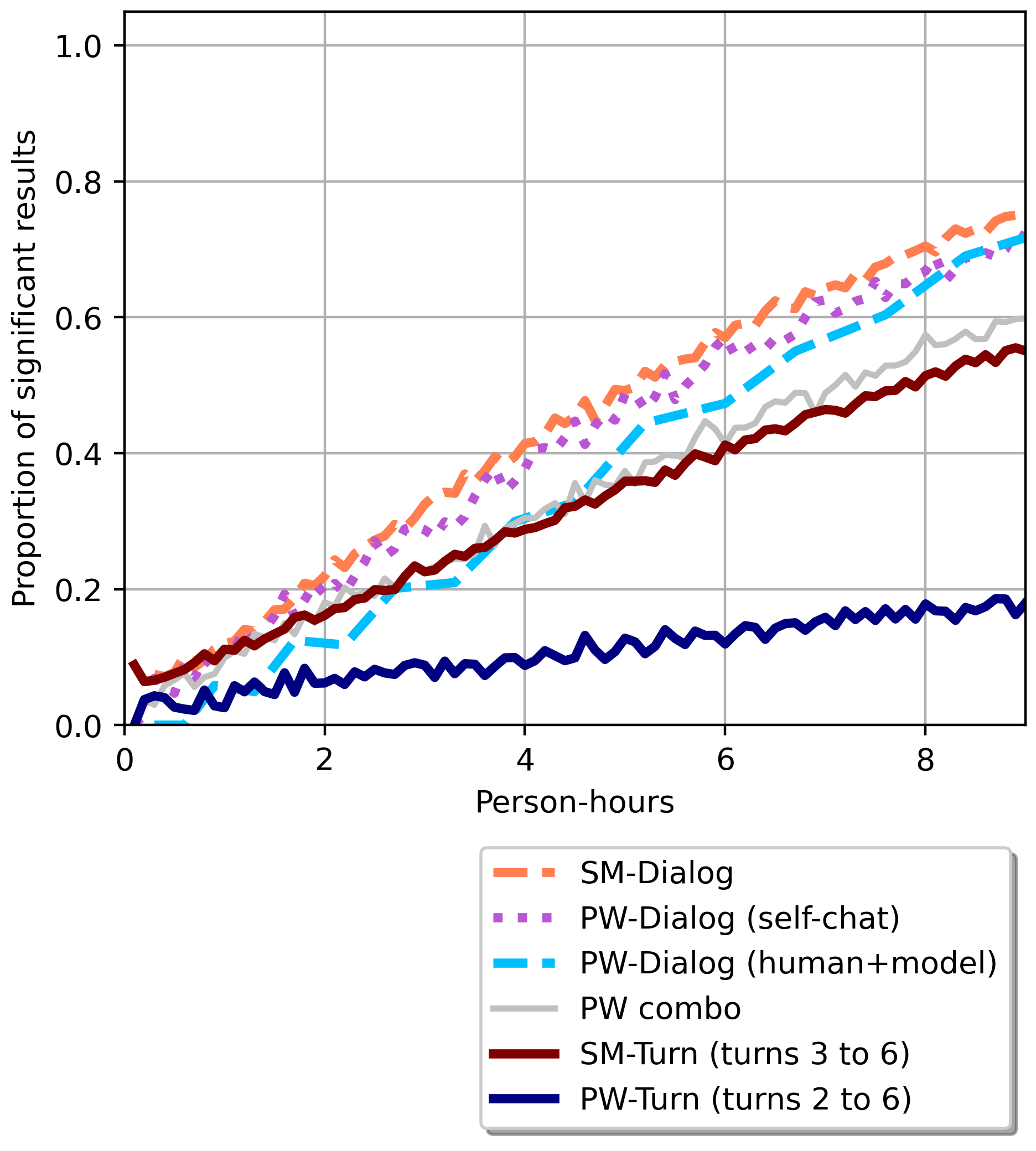}
  \caption{The time needed for statistical significance for the Size comparison between models (BlenderBot3B vs.\ BlenderBot90M).}
  \label{image:time_to_significance__size}
\end{figure}

\begin{figure}[h!]
\center
\includegraphics[width=\columnwidth]{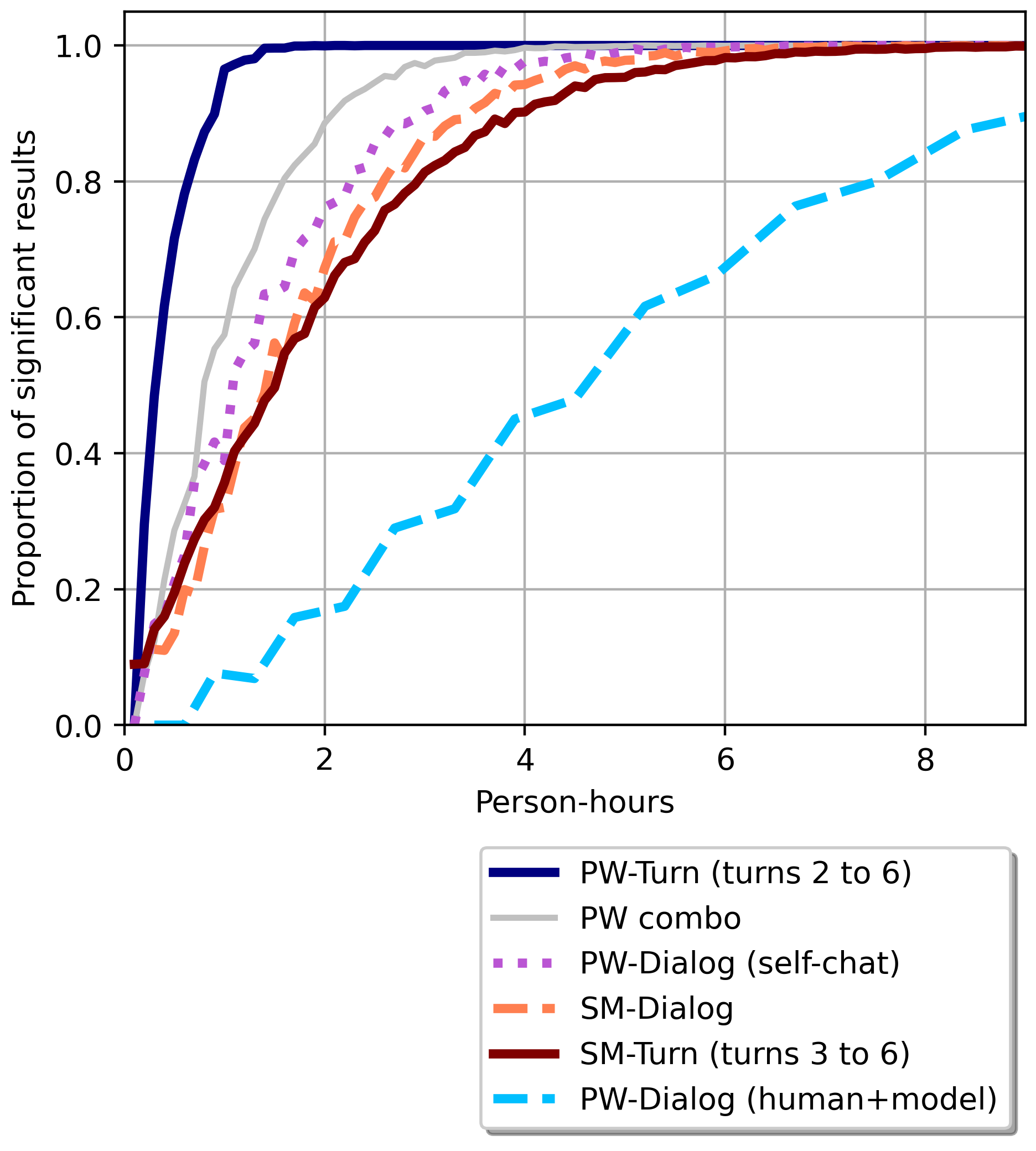}
  \caption{The time needed for statistical significance for the Fine-tuning comparison between models (BlenderBot3B vs.\ Reddit3B).}
  \label{image:time_to_significance__fine_tuning}
\end{figure}

Beyond win rates, another way to directly compare the relative usefulness of our various evaluation techniques is to estimate the amount of person-hours that must be spent on evaluations by crowdworkers in order to achieve a statistically significant result. These results (Figures~\ref{image:time_to_significance__length}, \ref{image:time_to_significance__size}, and~\ref{image:time_to_significance__fine_tuning}) roughly follow the patterns found by win rates (Section~\ref{sec:win_rates_all_techniques}). See Appendix~\ref{sec:time_to_sig_methodology} for a discussion of the assumptions made when producing these time estimates.

\section{Conclusion}

In this work we compare the extent to which different evaluation techniques are able to measure performance differences between dialogue models, and we show instances in which the performance varies between per-turn techniques and per-dialogue techniques, and between pairwise techniques and single-model techniques. A completely exhaustive analysis of the cases in which each technique is most appropriate would require measurement on many more pairs of models than the three studied here, and would likely require a dramatic scaling-up of labor for crowdworkers.

Nevertheless, the results shown here demonstrate the difficulty in anointing one evaluation technique as superior to all others regardless of the models being compared, and they suggest that a combination of techniques, or else a different technique entirely, may be necessary for optimal measurement of differences among models. A more universally ideal technique would likely need to investigate model performance per-turn but still be able to give an overall judgment of model quality across a conversation in order to capture elements of performance that manifest clearest in a single response vs.\ in aggregate. We demonstrate that combining evaluation scores from per-turn and per-dialogue techniques can bridge the gap in the performance differences between the two, but that this does not outperform either individual technique in all cases, at least in the way that we combined them.

Future improvements may also come from exploring other ways to amplify the weak signal from models with only slight performance differences such as BlenderBot3B and BlenderBot90M, perhaps by training workers to select responses based on general measures of conversational quality, as opposed to content that appeals to their personal interests. 
Improving sensitivity to roughly equivalent pairs of models such as these should in turn enable the comparison of models whose performance differences are smaller still.

While this work has concentrated on evaluating techniques that enable {\em differentiability} (one can differentiate between models) with efficiency (with less annotator hours), there are other desirable qualities as well. Some of these in particular are {\em diversity} of conversations \cite{hashimoto2019unifying}, {\em repeatability} of experiments, and {\em explainability} of results \cite{deriu2021survey}. While there is some discussion of the latter two topics in our experiments, these topics are fully deserving of a more thorough analysis than is provided here.




\bibliography{main}
\bibliographystyle{acl_natbib}

\newpage

\appendix

\section{Pairwise per-turn evaluation onboarding}
\label{sec:onboarding}

\begin{figure*}[t!]
\center
\includegraphics[width=13cm]{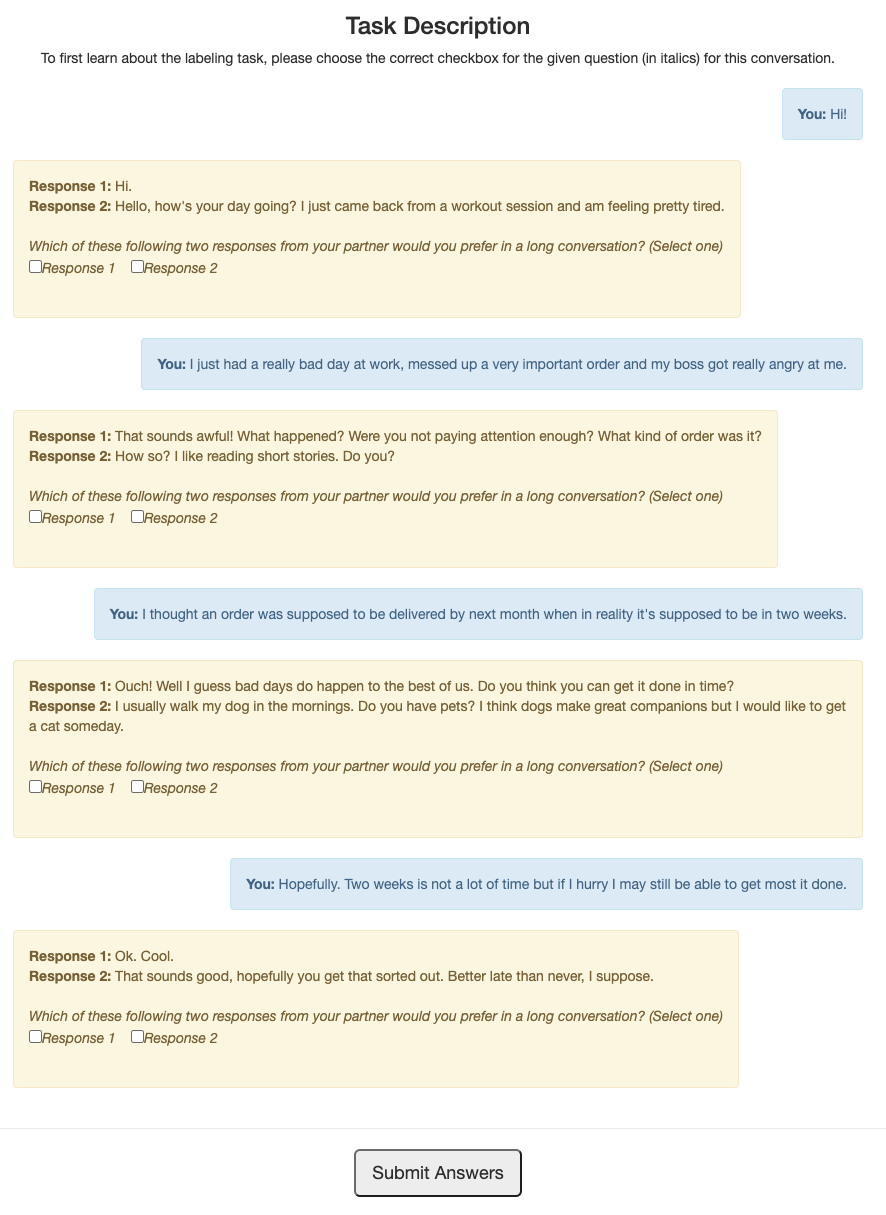}
  \caption{Screenshot of the onboarding process for crowdworkers for the \pwturn{} technique.}
  \label{image:onboarding}
\end{figure*}

In order to perform quality control on crowdworkers before the start of the conversation itself, we ask each worker to first annotate a conversation in which there are two possible responses for each turn of one of the speakers, one response of which is clearly better than the other (Figure~\ref{image:onboarding}). These pairs of responses vary slightly depending on which of the evaluation metrics is being tested. Workers must ultimately choose the correct response for all four pairs of responses but have two tries in which to do so.

\section{Checks used when filtering per-turn evaluations}
\label{sec:filtering_checks}
We check each conversation between a crowdworker and a Bot Speaker collected during \pwturn{} and \smturn{} evaluations against the criteria below to see if they have issues that warrant their exclusion from the final filtered set of evaluations. If at least one of the following problems are present, all evaluations from the crowdworker in question are filtered out of the results shown in this work:
\begin{itemize}
  \item The messages consist of less than three words on average
  \item The first message inputted by the worker contains a greeting (redundant, since a dummy \textit{``Hi!''}~message is already fixed to be the worker's first line of conversation) 
  \item Several of the messages are written using all capital letters
  \item Later messages are duplicates of the first one (i.e. the worker is repeating their messages throughout the conversation)
  \item One or more of the messages use offensive language
\end{itemize}

\section{Evaluation scores as a function of conversation turn}

\begin{table}[h!]
\centering
\begin{small}
\begin{tabular}{cccccccc}
\toprule
& & \multicolumn{6}{c}{\pwturn{}: turn index} \\
\cmidrule(lr){3-8}
Comp. & Metric & 1 & 2 & 3 & 4 & 5 & 6 \\
\midrule
Length & Pref & 40 & 56 & 67 & 67 & \textbf{74} & 72 \\
 & Human & 38 & \textbf{72} & 69 & 62 & \textbf{72} & 62 \\
 & Inter & 59 & 62 & 72 & \textbf{78} & 69 & 69 \\
\midrule
Size & Pref & 26 & 56 & 51 & 54 & 38 & \textbf{62} \\
 & Human & 29 & 58 & 50 & 46 & \textbf{67} & 58 \\
 & Inter & 31 & 59 & \textbf{62} & 52 & 45 & 45 \\
\midrule
FT & Pref & 70 & 78 & 74 & \textbf{89} & 81 & \textbf{89} \\
 & Human & 69 & 73 & \textbf{90} & 88 & 84 & 85 \\
 & Inter & 52 & 59 & \textbf{89} & 74 & 67 & 85 \\
\bottomrule  
\end{tabular}
\end{small}
\caption{Percentage win rates of BlenderBot3B vs.\ other models on \pwturn{} evaluations as a function of Bot Speaker turn. The highest win rate for each model comparison and evaluation metric is bolded. This is a tabular representation of the curves in Figure~\ref{image:win_rate_by_turn_idx}.}
\label{tab:win_rate_by_turn_idx}
\end{table}

\begin{table}[h!]
\centering
\begin{small}
\begin{tabular}{cccccccc}
\toprule
& & \multicolumn{6}{c}{\smturn{}: turn index} \\
\cmidrule(lr){3-8}
Model & Metric & 1 & 2 & 3 & 4 & 5 & 6 \\
\midrule
BB3B & Pref & 67 & 71 & \textbf{72} & 70 & \textbf{72} & 70 \\
 & Human & 70 & 63 & 72 & \textbf{76} & 71 & 70 \\
 & Inter & 42 & 43 & \textbf{47} & 43 & 45 & \textbf{47} \\
\midrule
BB3B-M0 & Pref & \textbf{74} & \textbf{74} & \textbf{74} & 67 & 66 & 72 \\
 & Human & 66 & \textbf{72} & 64 & 67 & 69 & 66 \\
 & Inter & 29 & \textbf{40} & 38 & 34 & 36 & 33 \\
\midrule
BB90M & Pref & \textbf{71} & 67 & 65 & 64 & 64 & 64 \\
 & Human & 59 & 64 & 61 & \textbf{70} & 65 & 67 \\
 & Inter & 39 & 35 & 39 & \textbf{41} & \textbf{41} & 38 \\
\midrule
Reddit3B & Pref & \textbf{67} & 63 & 49 & 50 & 54 & 46 \\
 & Human & \textbf{60} & 57 & 54 & 53 & 51 & 58 \\
 & Inter & 44 & 36 & 39 & \textbf{46} & 31 & 39 \\
\bottomrule  
\end{tabular}
\end{small}
\caption{Percentage success rates  of responses of various models on various evaluation questions (metrics) for \smturn{}, as a function of Bot Speaker turn. The highest  win rate turn for each model and evaluation metric is bolded.}
\label{tab:smturn_success_rate_by_turn_idx}
\end{table}

See Table~\ref{tab:win_rate_by_turn_idx} for the win rates of BlenderBot3B over other models using the \pwturn{} technique, as a function of Bot Speaker turn. (This is also shown as a line plot in Figure~\ref{image:win_rate_by_turn_idx}.) Table~\ref{tab:smturn_success_rate_by_turn_idx} likewise shows the success rates of model responses using the \smturn{} technique, as a function of Bot Speaker turn.

\section{Text justification for model response selection}
\label{sec:justification}

For \pwturn{} evaluations, we collect and analyze justification texts for each turn, after the worker selects a model response. We then group justification texts by model type and comparison.

To measure lengths of justifications, we split text strings into words (space-delimited), and we calculate the mean number of words in each sample. Results are shown in Table~\ref{tab:justification_lengths}.

\begin{table}[h!]
\centering
\begin{small}
\begin{tabular}{ccc}
\toprule
Comparison & Model & Avg. number of words \\
\midrule
Length & BB3B & 8.85 \\
 & BB3B-M0 & 7.70 \\
\midrule
Size & BB3B & 8.95 \\
 & BB90M & 8.81 \\
\midrule
Fine-tuning & BB3B & 9.40 \\
 & Reddit3B & 9.25 \\
\bottomrule  
\end{tabular}
\end{small}
\caption{Mean number of words in justifications given for BlenderBot3B vs.\ other models on \pwturn{} evaluations.}
\label{tab:justification_lengths}
\end{table}

For term importance, we use the \texttt{scikit-learn} \texttt{TfidfVectorizer} class to compute TF-IDF scores for each term in each model comparison. 

We use a list of English stopwords from the \texttt{NLTK} library to filter out common terms. Additionally, we discard terms that have a higher document frequency than 0.8.

The top 20 terms (descending order) for each model pairing are shown in Table~\ref{tab:justification_tfidf_terms}.

Our analysis reveals the following:

\begin{itemize}
  \item \textbf{Length comparison}: While it appears that many crowdworkers prefer longer responses overall, at least in some conversational turns some crowdworkers may prefer shorter responses.  The top terms in justifications for BlenderBot3B-M0 responses include ``simple'', ``short'' and ``direct'', while top terms in reasons for choosing BlenderBot3B include ``detailed'' and ``longer''. This shows that \pwturn{} evaluation does well in capturing sensitivity to length, and that workers' selections are due to their own preferences at a given conversational turn.
  
   Interestingly, in \pwturn{} we find that workers' justifications for choosing the BlenderBot3B-M0 responses are themselves on average shorter than for BlenderBot3B. Table~\ref{tab:justification_lengths} shows the mean justification lengths for different model pairings. The mean justification length for BlenderBot3B is 8.85 words, compared to a mean length of BlenderBot3B-M0 justifications of 7.7 words. This suggests that workers choosing shorter, ``simple'' responses may also be less detail-oriented.
  
  \item \textbf{Size comparison}: Top TF-IDF weighted terms from workers' justifications for both models contain a mix of references to the conversational content, such as ``hiking'', ``beach'' or ``dogs'', and conversational structure, such as ``relates'' or ``engaging''. By inspection, there are no discernible differences between these terms.
  
  \item \textbf{Fine-tuning comparison}: High TF-IDF-weighted terms in justifications given by workers who choose the BlenderBot3B model are mostly related to conversational flow, such as ``follows'', ``responds'', and ``acknowledges''. In contrast, terms appearing in justifications for the Reddit3B model are specific and often refer to the topic instead of conversational style, such as ``bath'', ``robot'', and ``paris''. This suggests that workers who choose the Reddit3B model often favor less natural responses because they contain particular references.
  
\end{itemize}

These nuanced differences are clear when evaluating model responses per turn, but are difficult to capture when evaluating the conversation as a whole. Analysis of worker justifications supports our hypothesis that differences in conversational quality are easier to identify in the \pwturn{} evaluation.

\begin{table*}[h!]
\centering
\begin{small}
\def\arraystretch{1.5}%
\begin{tabular}{ccp{11.75cm}}
\toprule
Comparison & Model & Top terms \\
\midrule
Length & BB3B & information, chosen, provides, follow, engaging, going, adds, speaker, detailed, interested, conversational, looks, little, \textit{play}, chat, new, \textit{pets}, includes, longer, \textit{tallies} \\
 & BB3B-M0 & \textit{day}, \textit{wow}, \textit{game}, going, simple, stays, short, direct, express, speaker, keeps, conversational, precise, \textit{western}, \textit{popular}, \textit{silk}, \textit{hands}, use, tone, elaborate \\
\midrule
Size & BB3B & message, information, easy, time, correct, want, interested, enjoy, change, relates, spend, prefer, fun, well, \textit{hiking}, \textit{pets}, go, moves, \textit{beach}, sound \\
 & BB90M & never, going, ok, excited, \textit{fav}, correct, changes, \textit{color}, new, engaging, personal, explain, \textit{ohio}, fluent, enjoy, \textit{hop}, \textit{hip}, listen, back, \textit{dogs} \\
\midrule
Fine-tuning & BB3B & follows, going, contradicts, great, responds, follow, responsive, never, contradict, flows, acknowledges, responses, responded, stays, looks, personal, keep, well, nothing, contradiction  \\
 & Reddit3B & \textit{bath}, personal, \textit{robot}, im, \textit{someone}, \textit{bubble}, detailed, flowing, \textit{play}, information, \textit{paris}, due, \textit{softball}, \textit{careers}, unique, direct, watch, told, \textit{book}, boring  \\
\bottomrule  
\end{tabular}
\end{small}
\caption{Top TF-IDF-weighted terms in justifications given for BlenderBot3B responses vs.\ other models on \pwturn{} evaluations. Terms that are irrelevant to conversational evaluation are italicized.}
\label{tab:justification_tfidf_terms}
\end{table*}

\section{Variability in win rate across evaluation techniques}
\label{sec:variability}

\begin{table}[h!]
\centering
\begin{small}
\begin{tabular}{lrrr}
\toprule
Technique & Length & Size & Fine-Tuning \\
\midrule
\pwturn{} & 10\% & 8\% & 11\% \\
\pwturn{} (WTA) & 18\% & 24\% & 13\% \\
\midrule
\pwdialog{} (self-chat) & 9\% & 9\% & 6\% \\
\midrule
\smturn{} & 14\% & 13\% & 12\% \\
\smturn{} (WTA) & 17\% & 16\% & 15\% \\
\midrule
\smdialog{} & 14\% & 15\% & 16\% \\
\bottomrule  
\end{tabular}
\end{small}
\caption{The variability of win rates of BlenderBot3B across different evaluation techniques, for different model comparisons (columns). Variability was measured by splitting each time-ordered set of ratings into chunks representing 45 minutes of crowdworker time each, and then computing the standard deviation of the win rate across chunks. Standard deviations are averaged across the three evaluation metrics (Section~\ref{sec:methods_pwpt}). Win rates for \pwturn{} were compiled over Bot Speaker turns 2 to 6 and for \smturn{} over turns 3 to 6, following Section~\ref{sec:win_rates_all_techniques}.}
\label{tab:variability_across_techniques}
\end{table}

Table~\ref{tab:variability_across_techniques} shows the variability in the win rates of BlenderBot3B per evaluation technique, as measured by splitting the ratings from each technique into chunks of equal crowdworker time. The win rates from \pwturn{}, \pwdialog{}, and \smturn{} vary least across chunks, largely because the mean time per rating is small, leading to a larger number of ratings per chunk and thus a more precise estimate obtainable within a given block of time.\footnote{We omit win rates of \pwdialog{} on conversations between a human and a model for simplicity. For this technique, the time to collect conversations varies non-linearly as a function of the number of ratings (Section~\ref{sec:time_to_sig_methodology}), and thus any dividing of ratings into chunks of equal crowdworker time would have to take this irregularly-spaced conversation collection time into account.} This suggests that calculating the per-conversation winner-takes-all win rate for the per-turn methods \pwturn{} and \smturn{} may be disadvantageous if having a precise measurement of the win rate is more important than one that is statistically significant.

\section{Methodology for calculations of the crowdsourcing time needed per model}
\label{sec:time_to_sig_methodology}

For the plots of the time to statistical significance in Section~\ref{sec:time_to_sig}, we consider ratings for each turn in Bot Speaker turns 2 through 6 for \pwturn{} and Bot Speaker turns 3 through 6 for \smturn{}, as in Section~\ref{sec:win_rates_all_techniques}.\footnote{We do not compute winner-takes-all scores for each conversation because in experiments this works less well. It 
greatly diminishes the total number of ratings per technique, and thus it increases the number of conversations needed to achieve statistical significance. We note that, compared to the per-dialogue technique \smdialog{}, the resulting rating is binary per evaluation metric in this case, which may contribute to poor performance.} We use a two-sided binomial test for \pwturn{} and \pwdialog{} and a two-sided independent $t$-test for \smturn{} and \smdialog{}. Significance is measured at a $p$-value of 5\%. When estimating the crowdsourcing time needed for each evaluation technique, we include an estimate of each technique's time to complete onboarding, which is mandatory before being approved to work on an evaluation.

For \pwdialog{} evaluations (i.e. Acute-Evals) on conversations between a human and a model, the labor costs involve collecting both conversations and rating pairs. This gives us a parameter to tune in this method: how many conversations to collect, and then how many times to reuse them when rating pairs of them. In our experiments, the number of conversations necessary is chosen such that each possible pairing of a conversation from one model and a conversation from another model should only be evaluated once at most: thus, if we have $N$ conversations for each of the two models being compared, we will be able to perform a maximum of $N^2$ \pwdialog{} evaluations on these conversations.\footnote{The potential drawback of this assumption is that the performance of the models will then likely be judged using only a relatively small handful of conversations, which may or may not be representative of the models' true performance.}

\section{Crowdsourcing time needed when combining single-model methods}
\label{sec:time_to_sig_single_model}

\begin{figure}[h!]
\center
\includegraphics[width=\columnwidth]{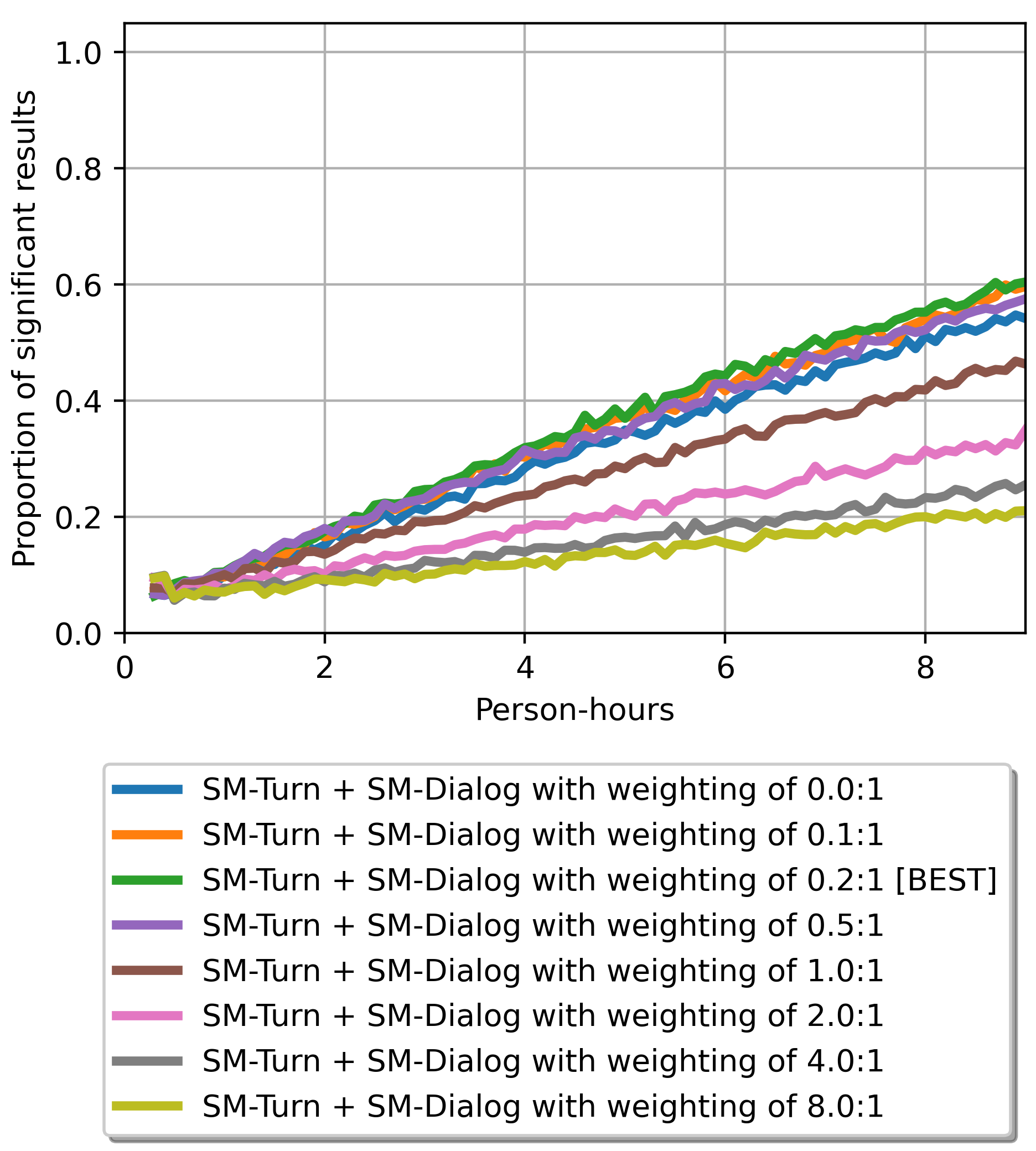}
  \caption{The time needed to measure a statistically significant result when averaging together per-conversation ratings of \smturn{} and \smdialog{} with the given weighting, for the Length comparison. The fastest weighting is marked with ``[BEST]''.}
  \label{image:time_to_sig__sm__length}
\end{figure}

\begin{figure}[h!]
\center
\includegraphics[width=\columnwidth]{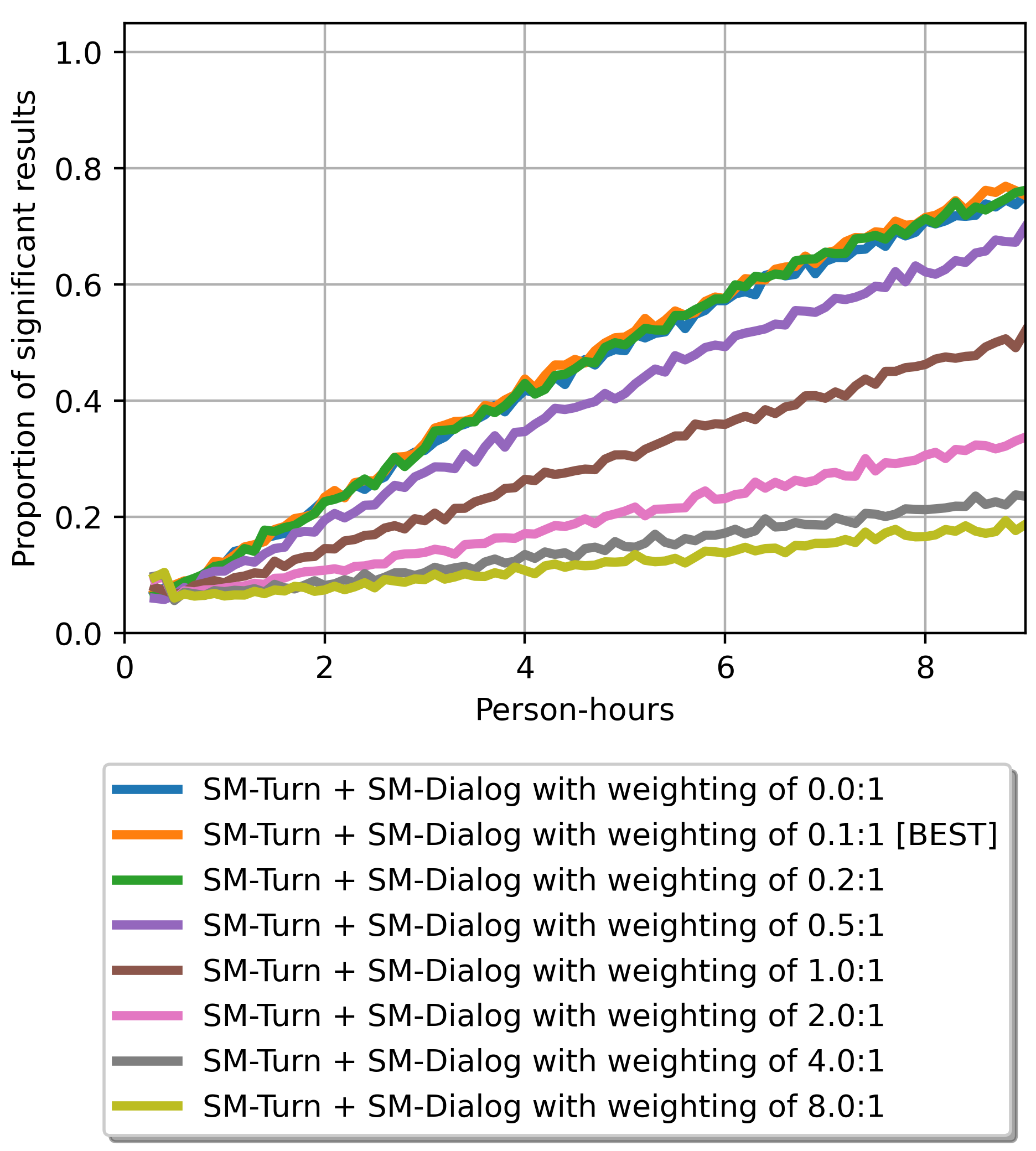}
  \caption{The time needed to measure a statistically significant result when averaging together per-conversation ratings of \smturn{} and \smdialog{} with the given weighting, for the Size comparison.}
  \label{image:time_to_sig__sm__size}
\end{figure}

\begin{figure}[h!]
\center
\includegraphics[width=\columnwidth]{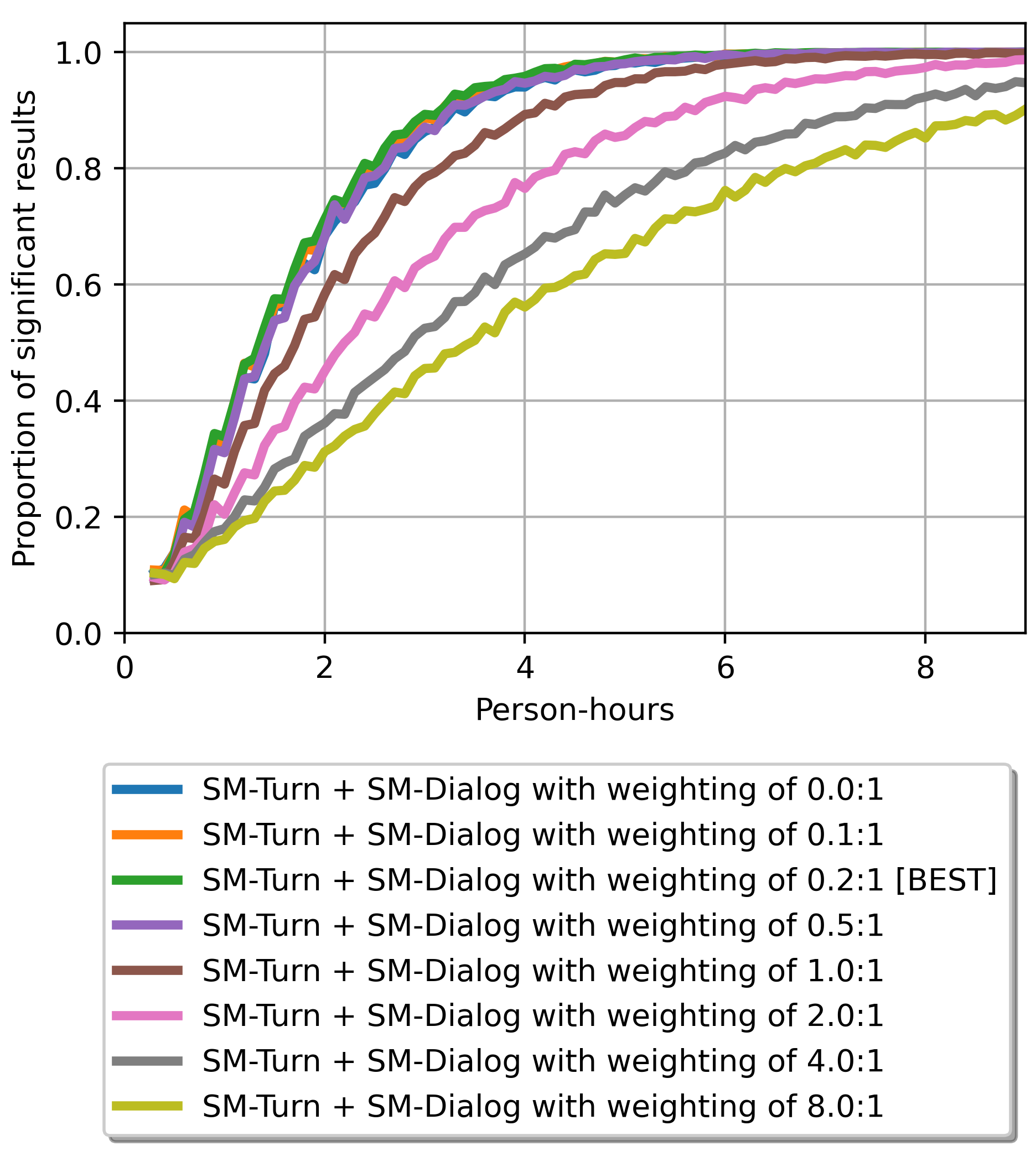}
  \caption{The time needed to measure a statistically significant result when averaging together per-conversation ratings of \smturn{} and \smdialog{} with the given weighting, for the Fine-tuning comparison.}
  \label{image:time_to_sig__sm__fine_tuning}
\end{figure}

Figures~\ref{image:time_to_sig__sm__length}, \ref{image:time_to_sig__sm__size}, and~\ref{image:time_to_sig__sm__fine_tuning} show the time needed to achieve a statistically significant difference between models when averaging together \smturn{} winner-takes-all success rates from Bot Speaker turns 3 to 6 (Section~\ref{sec:single_model_results}) with \smdialog{} Likert scores. To perform the weighted average between \smturn{} and \smdialog{} scores on each conversation, we first shift and scale the originally 1-to-5 \smdialog{} Likert scores to fall within the range $[0,1]$, matching the range of the individual binary \smturn{} success ratings. We see that statistical significance is reached fastest when weighting \smturn{} ratings much less heavily than \smdialog{} at a ratio of 1:5 or 1:10, which is to be expected given the already much stronger sensitivity of \smdialog{}.

\end{document}